# Mechanical Strength Prediction of Steel–Polypropylene Fiber-Based High-Performance Concrete Using Hybrid Machine Learning Algorithms

**Jagaran Chakma[a] , Zhiguang Zhou[a,*] , Badhan Chakma[b]**

[a] State Key Laboratory of Disaster Mitigation for Structures, Tongji University, Shanghai 200092, China

[b] School of Civil Engineering, Chongqing Jiaotong University, Chongqing 400074, China

---

## Abstract

This research develops and evaluates machine learning models to predict the mechanical properties of steel–polypropylene fiber-reinforced high-performance concrete (HPC). Three model families were investigated: Extra Trees with XGBoost (ET–XGB), Random Forest with LightGBM (RF–LGBM), and Transformer with XGBoost (Transformer–XGB). The target properties included compressive strength (CS), flexural strength (FS), and tensile strength (TS), based on an extensive dataset compiled from published experimental studies. Model training involved k-fold cross-validation, hyperparameter optimization, SHapley Additive exPlanations (SHAP), and uncertainty analysis to ensure both robustness and interpretability. Among the tested approaches, the ET–XGB model achieved the highest overall accuracy, with testing $R^2$ values of 0.994 for CS, 0.944 for FS, and 0.978 for TS, and exhibited the lowest uncertainty for CS and TS (approximately 13–16% and 30.4%, respectively). The RF–LGBM model provided the most stable and reliable predictions for FS ($R^2 = 0.977$), yielding the lowest uncertainty for FS (approximately 5–33%). The Transformer–XGB model demonstrated strong predictive capability ($R^2 = 0.978$ for TS and 0.967 for FS), but consistently showed the highest uncertainty, indicating reduced generalization reliability. SHAP analysis further indicated that fiber aspect ratios ($AR_1$ and $AR_2$), silica fume (Sfu), and steel fiber content (SF) were the most influential predictors of strength, whereas water content (W) and the water–binder ratio (W/B) consistently had negative effects. The findings confirm that machine learning models can provide accurate, interpretable, and generalizable predictions of HPC mechanical properties. These models offer valuable tools for optimizing concrete mix design and enhancing structural performance evaluation in engineering applications.

---

∗Corresponding author: Zhiguang Zhou,
E-mail addresses:
chakmajagaran@tongji.edu.cn (J. Chakma),
zgzhou@tongji.edu.cn (Z. Zhou),
badhanchakma737@gmail.com (B. Chakma)

## 1. Introduction

High-performance concrete (HPC) has garnered significant interest due to its exceptional mechanical properties, crack resistance, and long-term durability, particularly when reinforced with fibers such as steel and polypropylene fibers (PPF). While polypropylene fibers help to limit plastic shrinkage and improve ductility under tensile loads, steel fibers are known to greatly improve flexural strength and post-cracking tensile behavior [1], [2]. These fibers work together to create hybrid fiber-reinforced HPC, which is ideal for demanding structural applications as it allows for a more balanced enhancement in compressive strength (CS), flexural strength (FS), and tensile strength (TS) [3], [4].

Recent studies highlight the synergistic interaction between steel and polypropylene fibers in high-performance concrete. Through their high tensile strength and stiffness, steel fibers operate as micro-reinforcements, bridging cracks and facilitating the transfer of load after cracking [5], [6], [7]. They also improve flexural strength, energy absorption, and impact resistance. Low density and high elongation polypropylene fibers improve durability under tensile stress and minimize plastic shrinkage and microcracking during early hydration [8], [9], [10]. A more even distribution of stress and balanced enhancements in compressive, flexural, and tensile properties result from the incorporation of steel fibers to prevent the spread of macrocracks and polypropylene fibers to regulate the creation of microcracks [11]. This hybrid fiber method improves mechanical performance for applications requiring high load-bearing capacity and durability under challenging service factors, while also resolving brittleness in high-performance cementitious matrices [12], [13].

Accurately predicting the mechanical characteristics of fiber-reinforced high-performance concrete is essential for optimizing mix design and guaranteeing performance reliability [14], [15]. However, according to the nonlinear interactions among fiber type, amount, water-binder ratio, binder composition, and curing conditions, traditional empirical and regression-based models frequently fail to generalize across diverse mix designs or fiber combinations [16]. Consequently, academics have increasingly utilized machine learning (ML) as an effective instrument for modelling the intricate, multidimensional interactions that determine concrete performance [17], [18], [19].

Previous studies has employed decision tree-based techniques including random forest (RF), XGBoost (XGB), support vector regression (SVR), and artificial neural networks (ANN) to predict durability or compressive strength measures of plain or fiber-reinforced concrete [20], [21]. Although these models have demonstrated potential, when used on high-dimensional or diverse datasets, they frequently have restricted interpretability or limited predictive capability.

Recent advancements have explored hybrid models that integrate more expressive or feature-aware front-end models with the predictive power of boosting approaches such as XGBoost, LightGBM [22], [23], [24]. Specifically, enhanced the accuracy of their RAC strength prediction through the utilization of PSO-optimized GB and SVR models [25]. However, there is a paucity of comprehensive studies in the literature that implement advanced hybrid machine learning architectures in fiber-reinforced high-performance concrete, especially those employing intricate attention-based or tabular deep learning techniques [26], [27].

The present research proposes three innovative hybrid machine learning frameworks for predicting the CS, FS, and TS of steel and polypropylene fiber-reinforced HPC in order to close this gap:

1. Extra Tree + XGB: This method combines the Extra Tree algorithm, which increases variety through randomized feature splits, with XGBoost's gradient boosting to reduce residual prediction errors and improve robustness.

2. RF + LGBM: This model applies Random Forest's non-linear relationship capture and LightGBM's gradient boosting efficiency. For high-dimensional HPC datasets, the hybrid method balances interpretability, computational efficiency, and predicting accuracy.

3. Transformer + XGB: This method uses Transformer's self-attention mechanism to record complicated feature interactions, followed by XGBoost regression for precise predictions.

In order to ensure model robustness and generalizability, these hybrid models will be further improved by utilizing random search optimization methods with k-fold cross-validation. SHapley Additive exPlanations (SHAP) will be integrated to determine the most significant input features influencing the predictions in order to get beyond the interpretability limitation of ensemble and comprehensive models.

Moreover, given that the proposed hybrid models are innovative, the study will explore about using uncertainty quantification techniques to assess the predicted accuracy and model reliability—two important factors for engineering decision-making in the actual world.

## 2. Methodology

This research used a systematic approach to develop, evaluate, and analyze hybrid machine learning models for predicting the mechanical characteristics of fiber-reinforced high-performance concrete (HPC). The chosen methodology is shown in Fig.1 and described in detail below.

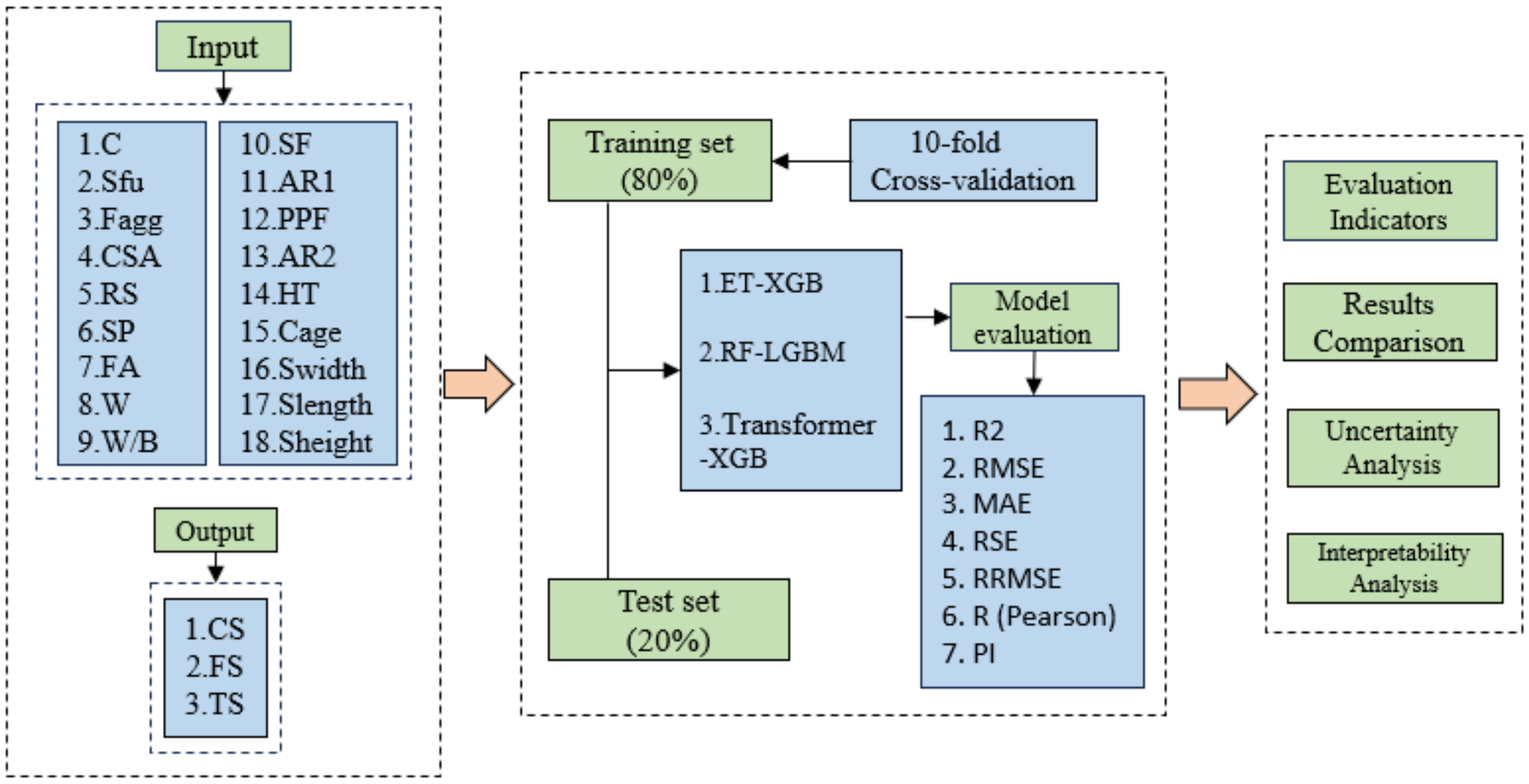

Fig. 1 The flowchart of the methodology

## 2.1 Statistical analysis of data description

This study analyzes data derived from earlier experimental investigations to predict HPC mechanical characteristics using hybrid machine learning algorithms. The mix compositions and strength characteristics of HPC are among the data taken from previous studies [3], [28], [29], [30], [31], [32], [33], [34], [35], [36], [37], [38], [39], [40], [41], [42], [43], [44], [45]. There are 302 data for tensile strength, 386 data for flexural strength, and 703 data for compressive strength in the dataset. To ensure accurate model validation, these data were divided into training and

Table 1 Statistical analysis of the CS dataset.

| Statistics | C (kg/m3) | Sfu (kg/m3) | Fagg (kg/m3) | CSA (kg/m3) | RS (kg/m3) | SP (kg/m3) | FA (kg/m3) | W (kg/m3) | W/B | SF (%) | AR1 | PPF (%) | AR2 | HT (℃) | Cage (Day) | Swidth ($cm^3$) | Slength ($cm^3$) | Sheight ($cm^3$) | CS (MPa) |
|---|---|---|---|---|---|---|---|---|---|---|---|---|---|---|---|---|---|---|---|
| Mean | 453.18 | 21.32 | 646.09 | 849.82 | 676.71 | 6.18 | 18.73 | 184.01 | 0.36 | 0.73 | 37.84 | 0.45 | 8.43 | 16.01 | 26.21 | 11.79 | 11.79 | 11.79 | 55.31 |
| STD | 125.99 | 29.61 | 266.32 | 291.72 | 216.91 | 6.67 | 41.74 | 182.36 | 0.07 | 1.45 | 44.26 | 0.76 | 12.35 | 9.18 | 19.56 | 3.08 | 3.08 | 3.08 | 45.82 |
| Min | 180 | 0 | 0 | 0 | 0 | 0 | 0 | 125 | 0.14 | 0 | 0 | 0 | 0 | 0 | 1 | 0 | 0 | 0 | 18.39 |
| 25% | 405 | 0 | 617.75 | 800 | 628 | 0.93 | 0 | 156 | 0.3 | 0 | 0 | 0 | 3.5 | 20 | 7 | 10 | 10 | 10 | 38 |
| 50% | 450 | 0 | 712 | 898 | 724 | 4.2 | 0 | 172 | 0.35 | 0 | 40 | 0.15 | 6 | 20 | 28 | 10 | 10 | 10 | 53.38 |
| 75% | 500 | 48 | 778.75 | 1080 | 800 | 9.9 | 0 | 186 | 0.42 | 0.9 | 64 | 0.57 | 8.33 | 23 | 28 | 15 | 15 | 15 | 70.8 |
| Max | 980 | 120 | 1250 | 1200 | 980 | 35 | 135 | 805 | 0.55 | 14.18 | 250 | 4 | 60 | 23 | 91 | 30 | 30 | 30 | 138.75 |

Note: STD is the standard deviation; “25%”, “50%” and “75%” in the table are the 25th, 50th (median), and 75th percentiles;

Table 2 Statistical analysis of the FS dataset.

| Statistics | C (kg/m3) | Sfu (kg/m3) | Fagg (kg/m3) | CSA (kg/m3) | RS (kg/m3) | SP (kg/m3) | FA (kg/m3) | W (kg/m3) | W/B | SF (%) | AR1 | PPF (%) | AR2 | HT (℃) | Cage (Day) | Swidth ($cm^3$) | Slength ($cm^3$) | Sheight ($cm^3$) | FS (MPa) |
|---|---|---|---|---|---|---|---|---|---|---|---|---|---|---|---|---|---|---|---|
| Mean | 468.28 | 16.09 | 635.47 | 918.89 | 703.98 | 6.83 | 19.37 | 178.03 | 0.36 | 0.92 | 40.67 | 0.66 | 5.59 | 18.08 | 24.72 | 9.12 | 40.01 | 9.37 | 9.48 |
| STD | 134.92 | 26.07 | 290.61 | 389.03 | 214.21 | 6.36 | 40.53 | 42.65 | 0.07 | 1.70 | 49.99 | 0.94 | 5.81 | 7.74 | 16.89 | 3.84 | 16.11 | 3.79 | 14.73 |
| Min | 200 | 0 | 0 | 0 | 0 | 0 | 0 | 125 | 0.25 | 0 | 0 | 0 | 0 | 0 | 1 | 0 | 0 | 0 | 1 |
| 25% | 420 | 0 | 617 | 800 | 628 | 2 | 2 | 158 | 0.3 | 0 | 0 | 0 | 0.55 | 20 | 7 | 8 | 40 | 10 | 5.03 |
| 50% | 450 | 0 | 712 | 939 | 729 | 5.16 | 0 | 175 | 0.35 | 0 | 42 | 0.38 | 4 | 20 | 28 | 10 | 40 | 10 | 6.5 |
| 75% | 500 | 40 | 823 | 1110 | 823 | 12.5 | 0 | 186 | 0.43 | 1 | 64 | 0.75 | 8.3 | 23 | 28 | 10 | 50 | 10 | 9.24 |
| Max | 980 | 120 | 1250 | 1200 | 980 | 18.68 | 140 | 372.4 | 0.55 | 14.18 | 250 | 4 | 24.19 | 23 | 91 | 15 | 72 | 15 | 14.9 |

Table 3 Statistical analysis of the TS dataset.

| Statistics | C (kg/m3) | Sfu (kg/m3) | Fagg (kg/m3) | CSA (kg/m3) | RS (kg/m3) | SP (kg/m3) | FA (kg/m3) | W (kg/m3) | W/B | SF (%) | AR1 | PPF (%) | AR2 | HT (℃) | Cage (Day) | Sdiameter ($cm^3$) | Sheight ($cm^3$) | TS (MPa) |
|---|---|---|---|---|---|---|---|---|---|---|---|---|---|---|---|---|---|---|
| Mean | 431.19 | 12.13 | 700.40 | 868.22 | 693.87 | 5.27 | 27.42 | 181.42 | 0.39 | 0.73 | 41.20 | 0.67 | 5.27 | 16.63 | 24.83 | 7.89 | 16.28 | 4.99 |
| STD | 122.03 | 20.39 | 143.11 | 253.13 | 150.99 | 6.13 | 52.33 | 19.64 | 0.07 | 1.27 | 32.78 | 1.00 | 4.53 | 8.57 | 16.49 | 5.74 | 11.86 | 8.69 |
| Min | 180 | 0 | 0 | 0 | 0 | 0 | 0 | 156 | 0.25 | 0 | 0 | 0 | 0 | 0 | 1 | 0 | 0 | 0.39 |
| 25% | 417 | 0 | 620 | 758.88 | 620 | 0 | 0 | 163.8 | 0.34 | 0 | 0 | 0 | 3.29 | 20 | 14 | 0 | 0 | 2.7 |
| 50% | 446 | 0 | 712 | 875 | 712 | 4.2 | 0 | 177.1 | 0.4 | 0 | 46.67 | 0.25 | 5.31 | 20 | 28 | 10 | 20 | 3.74 |
| 75% | 468 | 26 | 823 | 1080 | 823 | 5.48 | 0 | 198 | 0.45 | 1 | 80 | 0.9 | 6 | 22 | 28 | 10 | 30 | 4.94 |
| Max | 800 | 65 | 960 | 1200 | 980 | 18.68 | 140 | 242 | 0.48 | 6.1 | 80 | 4 | 24.19 | 23 | 91 | 15 | 30 | 9.2 |

testing sets in an 8:2 ratio. Tables 1–3 provide a summary of input and output factors that describe compressive, flexural, and tensile strengths.

The variables that are utilized for prediction contain Cement (C), Silica fume (Sfu), Fine aggregate (Fagg), Crushed Stone Aggreagate (CSA), River Sand (RS), Super plasticizer (SP), Fly Ash (FA), Water (W), Water to binder ratio (W/B), Steel fiber content (%), Aspect ratio (AR1), Polypropylene fiber (PPF), Aspect ratio (AR2), Hydration temperature (HT), Curing age (Cage), Specimen width (Swidth), Specimen length (Slength), Specimen height (Sheight), Specimen diameter (Sdiameter). Statistical summaries of the input and target variables for the flexural and tensile strength datasets are presented separately. These summaries include data on essential percentiles ($P_{25}$, $P_{50}$, and $P_{75}$), mean, standard deviation, minimum, and maximum values.

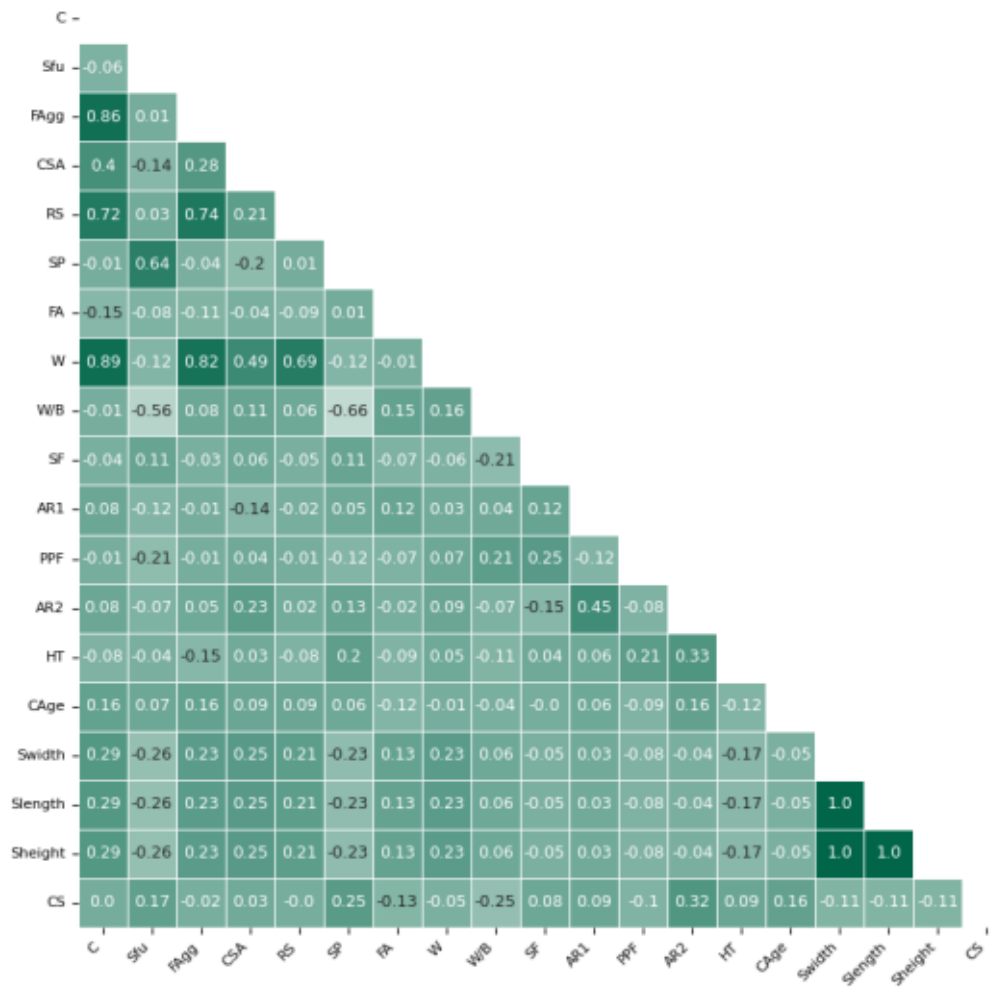

1(a)

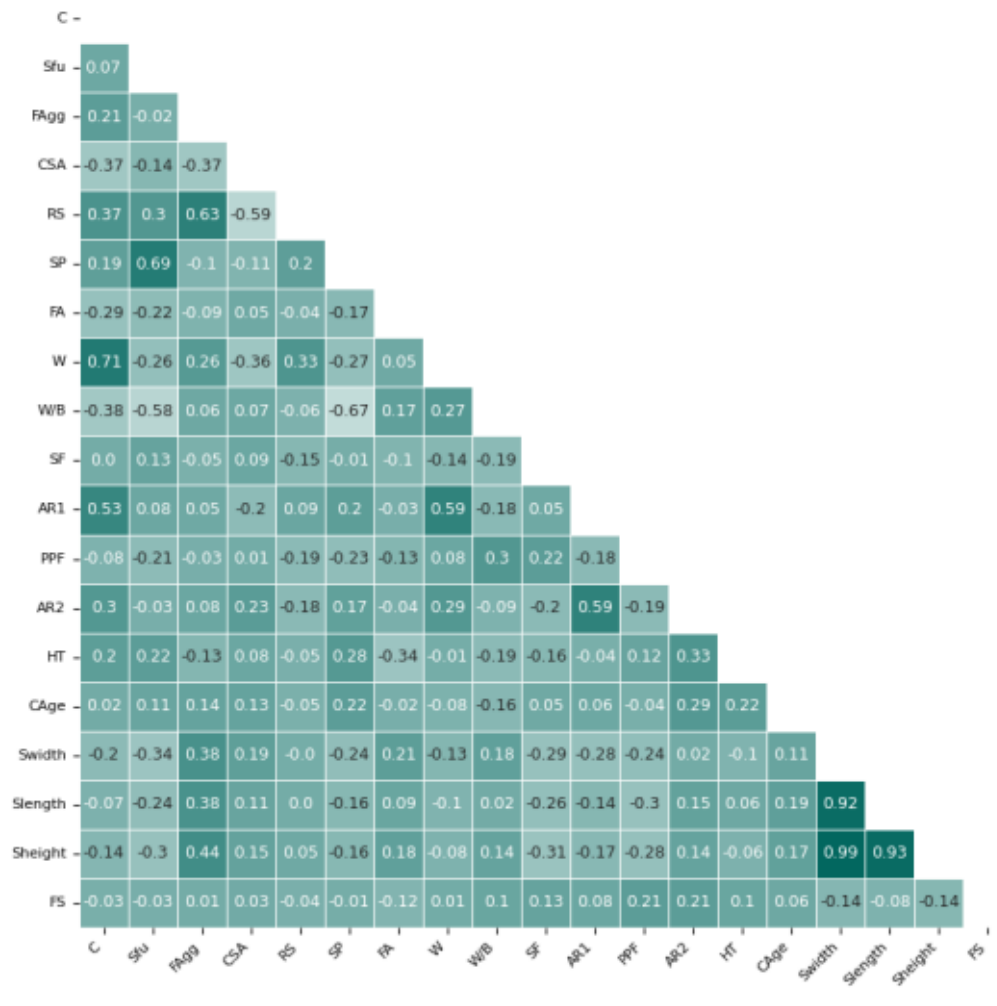

1(b)

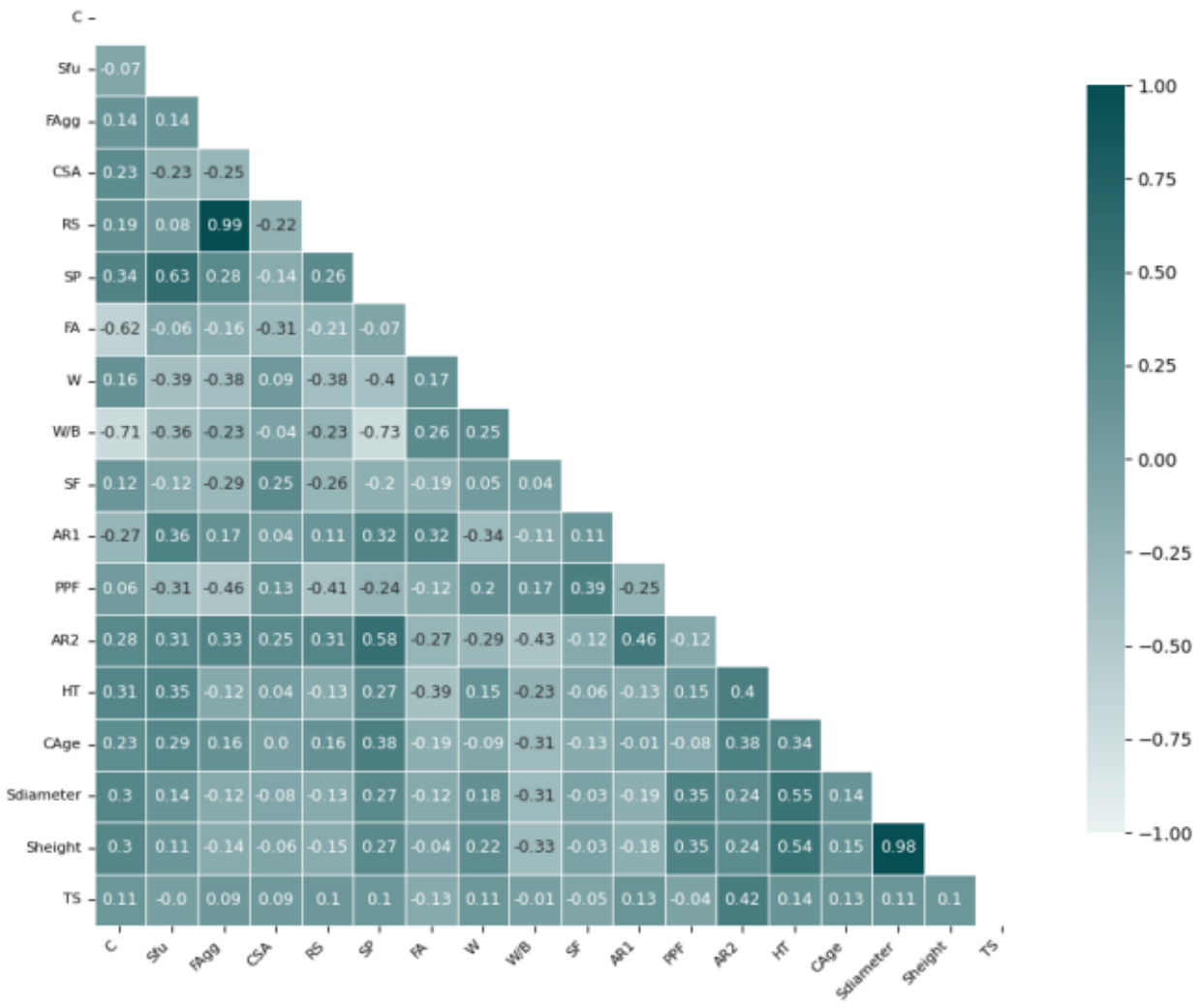

1(c)

Fig. 1. The Pearson correlation coefficient matrix of features: (a) CS, (b) FS, (c) TS.

The Pearson correlation matrices Figures 1 (a), (b), and (c), demonstrate clear correlations between parameters and their strengths. For Compressive strength, AR2, SP, and Sfu exhibit the most substantial positive connections (about 0.32, 0.25, 0.17), while W/B and FA demonstrate the most significant negative correlations (around −0.25, −0.13); curing age exerts a small influence (approximately 0.16). For Flexural strength, PPF and AR2 are the most influential factors (≈0.21), followed by SF (≈0.13), with negative correlations observed in specimen width, height, and FA (≈−0.14 to −0.12). In Tensile strength, AR2 is the predominant factor (≈0.42), while HT, CAge, and AR1 exhibit moderate positive correlations (≈0.13–0.14), and SF and PPF show modest negative correlations (≈−0.05, −0.04). The geometric dimensions exhibit a high degree of intercorrelation (~0.99) due to following the standardized testing of specimen sizes. In summary, binder and water parameters mostly influence CS and FS, whereas TS is more responsive to fiber-curing interactions, therefore preferring nonlinear ML approaches.

## 2.2 Machine Learning Models and Evaluation Metrics

### 2.2.1 Hybrid Machine Learning Frameworks

To precisely assess the mechanical properties of hybrid fiber-reinforced high-performance concrete (HPC), it is crucial to consider intricate nonlinearities, multi-scale interactions, and varied feature contributions. Traditional machine learning methods that rely on a singular model often exhibit inadequate generalization or insufficient interpretability. During this research, three novel hybrid frameworks were developed to address these challenges. The frameworks combined ensemble learning methodologies and attention-based structures. To enhance the predicted accuracy, robustness, and interpretability of compressive strength (CS), flexural strength (FS), and tensile strength (TS) predictions, these frameworks utilize the synergies provided by several algorithms.

### 2.2.2 Extra Trees and XGBoost (ET–XGB)

The ET–XGB framework integrates the Extra Tree (ET) algorithm with XGBoost (XGB) to use their complementary advantages. Extra Tree generates randomized decision trees by randomly selecting both features and thresholds, hence enhancing diversity and mitigating overfitting in comparison to deterministic tree-based models. This unpredictability enhances feature diversity, which is especially advantageous for HPC datasets characterized by variable material properties and mix proportions. Following to ET's feature splitting, XGB executes gradient boosting on the resultant residuals to enhance predictions and mitigate bias, utilizing a regularized objective function:

$$\hat{y} = \sum_{\mathrm{k}=1}^{K} f_{\mathrm{k}}(x), \quad f_{\mathrm{k}} \in \mathcal{F} \tag{1}$$

$$\mathcal{L} = \sum_{i=1}^{n} l(y_i, \hat{y}i) + \sum \mathrm{k} = 1^{K}\left(\gamma T + \frac{1}{2}\lambda\|\omega\|^2\right) \tag{2}$$

Where $T$ represents the number of leaves in each boosted tree, while $\gamma$ and $\lambda$ are regularization coefficients that govern model complexity.

### 2.2.3 Random Forest and LightGBM (RF–LGBM)

The RF–LGBM framework combines Random Forest (RF) with LightGBM (LGBM) to harmonize nonlinear modeling capabilities, computational efficiency, and interpretability. Random Forest functions by integrating predictions from multiple decision trees, which are trained on bootstrap samples via random feature selection, expressed as:

$$\hat{y} = \frac{1}{M}\sum_{\mathrm{m}=1}^{M} f_{\mathrm{m}}(x) \tag{3}$$

where $M$ represent the quantity of trees. This ensemble method significantly diminishes variation and elucidates complex connections among input factors. LightGBM enhances this by utilizing histogram-based gradient boosting, incorporating gradient-based one-side sampling (GOSS) and exclusive feature bundling (EFB) to expedite computation while maintaining accuracy. This hybridization is especially beneficial for high-dimensional HPC data where computational expense is a consideration.

2.2.4 Transformer and XGBoost (Transformer–XGB)

The Transformer–XGB framework integrates the Transformer architecture, known for its self-attention mechanism, with the boosting prowess of XGBoost. The Transformer module captures long-range interdependence and complex inter-feature interactions through the computation of scaled dot-product attention.

$$Attention(Q, K, V) = softmax\left(\frac{QK^T}{\sqrt{d_{\mathrm{k}}}}\right)V \tag{4}$$

In this context, $Q$, $K$, and $V$ denote the query, key, and value matrices obtained from input characteristics, whereas $d_k$ signifies the dimensionality scaling factor. The attention module's output encodes globally contextualized feature representations, which are then processed by XGBoost regression to generate final strength predictions. This hybrid methodology utilizes the interpretability of tree models alongside the representational capabilities of deep attention processes, providing a considerable benefit for intricate HPC combinations.

2.2.5 Evaluation metrics

To assess the predictive performance of the hybrid machine learning models, seven statistical metrics were utilised: mean absolute error (MAE), root mean square error (RMSE), coefficient of determination (R^2), relative squared error (RSE), relative root mean square error (RRMSE), Pearson correlation coefficient (R), and performance index (PI). Here formulas are presented as follows:

$$R^2 = 1 - \frac{\sum_{i=1}^{n}(y_i - \hat{y}_i)^2}{\sum i=1^n (y_i - \bar{y})^2} \tag{5}$$

$$RMSE = \sqrt{\frac{1}{n}\sum_{i=1}^{n}(y_i - \hat{y}_i)^2} \tag{6}$$

$$MAE = \frac{1}{n}\sum_{i=1}^{n}|y_i - \hat{y}_i| \tag{7}$$

$$RSE = \frac{\sum_{i=1}^{n}(y_i - \hat{y}_i)^2}{\sum i=1^n (y_i - \bar{y})^2} \tag{8}$$

$$RRMSE = \frac{\sqrt{\frac{1}{n}\sum_{i=1}^{n}(y_i - \hat{y}_i)^2}}{\bar{y}} \times 100\% \tag{10}$$

$$R = \frac{\sum_{i=1}^{n}(y_i - \bar{y})(\hat{y}i - \hat{y})}{\sqrt{\sum i=1^n (y_i - \bar{y})^2 \sum_{i=1}^{n}(\hat{y}_i - \hat{y})^2}} \tag{11}$$

$$PI = \frac{R^2}{1+RRMSE} \tag{12}$$

where n represents the total number of samples, $y_i$ and $\hat{y}_i$ signify the actual and predicted values, respectively, and $\bar{y}$ denotes the mean of the actual values.

### 2.3 SHAP analysis of the algorithms

A SHAP application, which denotes SHapley Additive exPlanations, was employed to elucidate the contribution of input variables to the predictions generated by hybrid models. The SHAP method, grounded in the principles of cooperative games, assigns an importance rating to each feature by assessing the marginal contribution of each feature across all possible feature combinations. This guarantees that the prediction "credit" is allocated among the input features in a manner that is equitable and uniform. SHAP, unlike conventional sensitivity analyses, provides both local interpretability, which involves elucidating individual predictions, and global insights into the impact of variables over the entire dataset. Due to its additive nature, it facilitates direct comparison of feature impacts, rendering it especially suitable for complex models like as Transformer–XGB and tree-based ensembles. This study enhances the transparency of the prediction framework by identifying the key parameters that affect the mechanical properties of hybrid fiber-reinforced high-performance composites (HPC) using SHAP.

### 2.4 Uncertainty assessment of hybrid algorithms

An uncertainty index was utilized to assess the resilience and dependability of the hybrid models (ET–XGB, RF–LGBM, and Transformer–XGB) by incorporating both prediction accuracy and variability. The conventional uncertainty metric at a 95% confidence interval is expressed as:

$$U = z \cdot \sqrt{(RMSE^2 + \sigma^2)} \tag{13}$$

In this context, RMSE denotes the mean prediction error, σ signifies the standard deviation of prediction errors (variability), and z = 1.96 is associated with the 95% confidence level assuming a normal distribution. This formulation provides a clear interpretation by integrating mean error and dispersion into a singular value.

However, this method does not consider differences in scale between target variables, like the difference between compressive and tensile strength. To address this limitation, a better normalized measure was used:

$$U_{norm} = \frac{\sqrt{(RMSE^2 + \sigma^2)}}{\bar{y}} \times 100\% \quad (14)$$

where ȳ represents the mean of the observed values. This normalisation quantifies uncertainty as a percentage, facilitating equitable comparisons among various mechanical properties with differing magnitudes. A diminished $U_{norm}$ signifies enhanced prediction reliability in relation to the property's scale.

The simultaneous use of both metrics offers thorough insight: Eq. (13) measures absolute uncertainty, whereas Eq. (14) analyses relative uncertainty across many outputs. This integrated technique provides a more resilient framework for assessing prediction stability in hybrid fiber-reinforced high-performance concrete models, particularly over varying property scales.

## 3. Results and discussion

### 3.1 Extra Tree-XGB model

The Extra Tree–XGB (ET-XGB) hybrid model was developed by integrating the Extra Tree (ET) algorithm with the Extreme Gradient Boosting (XGB) framework to improve its accuracy in predicting of the material's mechanical properties, including compressive strength (CS), flexural strength (FS), and tensile strength (TS). The optimal hyperparameter values acquired after optimization are listed in Table 4. The ET component was configured with 500 estimators, a minimum sample split of 2, and a minimum sample leaf of 1. The XGB component was optimized with a learning rate of 0.005, a maximum depth of 5, and subsample and colsample by tree ratios of 0.4, among other parameters. The optimized settings enabled the model to balance the bias–variance trade-offs while reducing the risk of overfitting.

Table 4. The optimal values of the ET-XGB model for CS, FS, and TS.

| **Hybrid algorithms** | | **Hyperparameter optimization** | **Optimal values** |
|---|---|---|---|
| (1) **Extra Tree - XGB** | **ET** | No. of estimators | 500 |
| | | Min sample split | 2 |
| | | Min sample leaf | 1 |
| | | No of jobs | -1 |
| | | Max depth | None |
| | | Random state | 100 |
| | **XGB** | Reg alpha | 5.0 |
| | | Reg lamda | 10.0 |
| | | Learning rate | 0.005 |
| | | Min child weight | 10 |
| | | Max depth | 5 |
| | | Gamma | 1.0 |
| | | Subsample | 0.4 |
| | | Colsample by tree | 0.4 |
| | | Random state | 100 |
| | | No. of estimators | 500 |

Table 5. Evaluation metrics of the ET-XGB model for CS, FS, and TS.

| ET-XGB | Training set | | | Testing set | | |
|---|---|---|---|---|---|---|
| | CS | FS | TS | CS | FS | TS |
| **$R^2$** | 0.995 | 0.994 | 0.999 | 0.994 | 0.944 | 0.978 |
| **RMSE** | 2.609 | 0.987 | 0.115 | 5.115 | 4.842 | 0.999 |
| **MAE** | 1.144 | 0.330 | 0.047 | 3.051 | 1.823 | 0.467 |
| **RSE** | 0.005 | 0.006 | 0.002 | 0.005 | 0.056 | 0.021 |
| **RRMSE** | 0.049 | 0.109 | 0.023 | 0.083 | 0.435 | 0.218 |
| **R (Pearson)** | 0.998 | 0.998 | 0.999 | 0.998 | 0.990 | 0.996 |
| **PI** | 0.024 | 0.055 | 0.011 | 0.042 | 0.219 | 0.109 |

The predictive performance of the ET-XGB model was evaluated using multiple statistical metrics, with the findings contained in Table 5. On the training set, the model achieved very high $R^2$ values of 0.995, 0.994, and 0.999 for CS, FS, and TS, respectively. Consequently, low RMSE and MAE values were observed, proving the model's superior fit to the training data. The Pearson correlation coefficients (R) exceeded 0.99 for all three features, hence validating the model's resilience.

On the testing set, the ET-XGB model also demonstrated strong generalization performance. The $R^2$ values observed were 0.994 for CS, 0.944 for FS, and 0.978 for TS, indicating that the model effectively explained the majority of the variation in the experimental data. The prediction errors, indicated by RMSE and MAE values, stayed within acceptable limits for all attributes, while FS exhibited slightly higher values than CS and TS. The findings indicate that the model is more adept at capturing the nonlinear relationships that govern compressive and tensile strength in comparison to flexural strength.

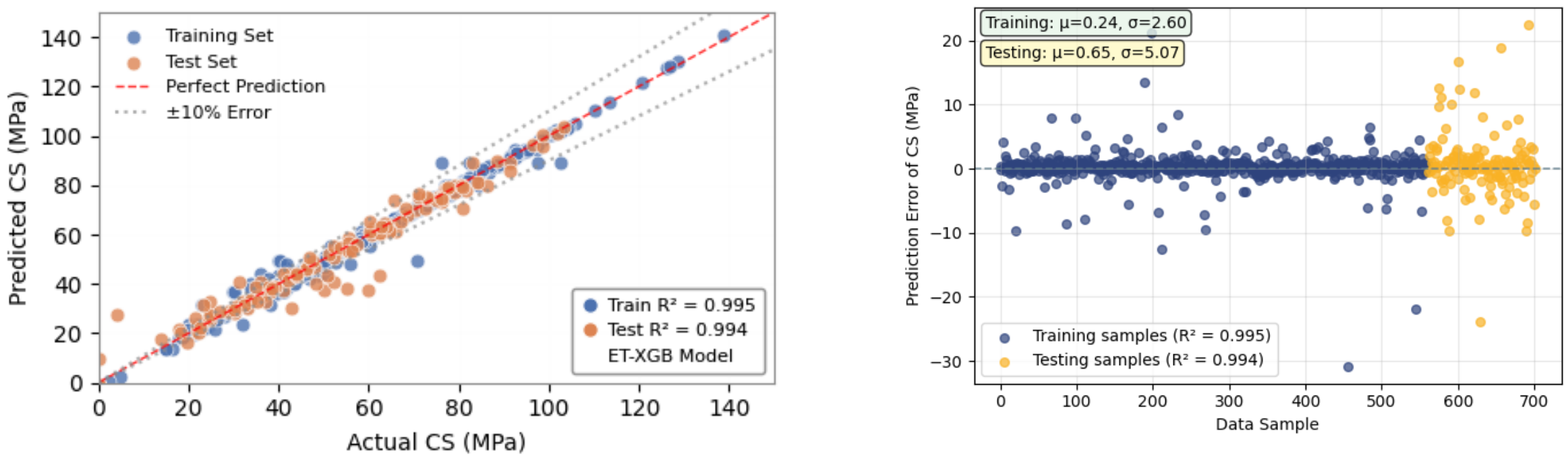

2(a)

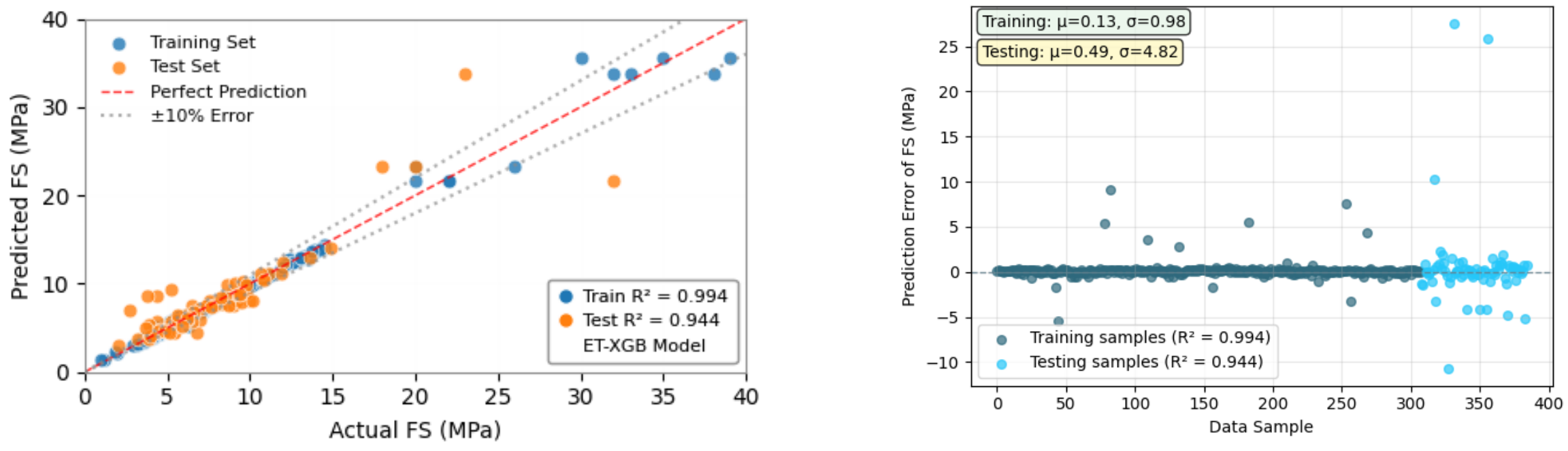

2(b)

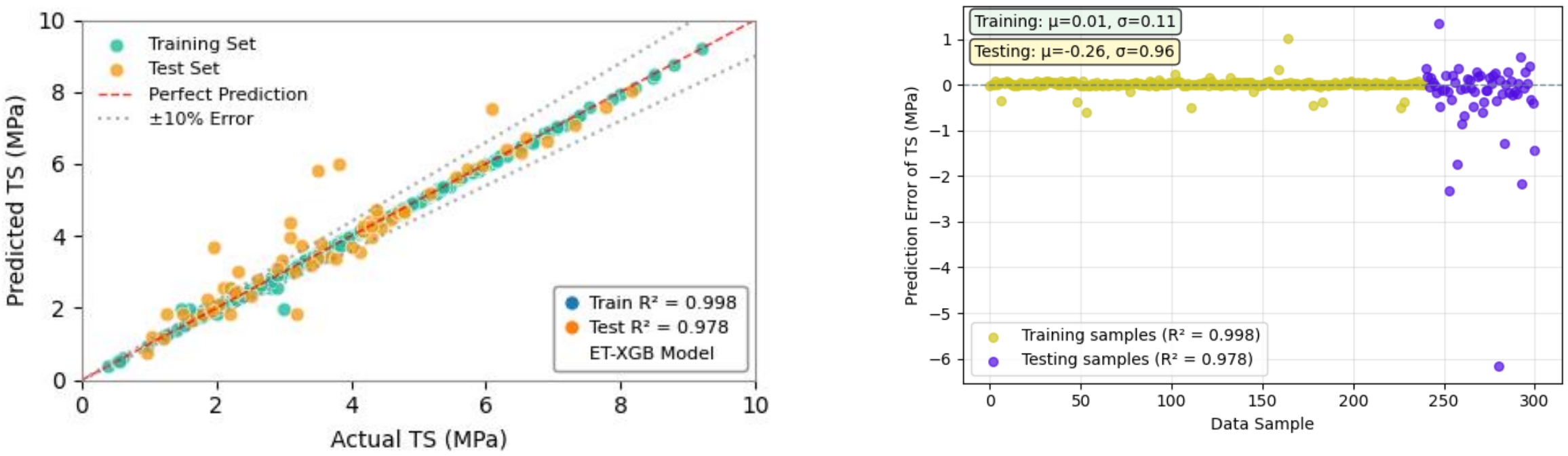

2(c)

Fig. 2. illustrates the correlation with actual and predicted results for (a) CS, (b) FS, and (c) TS derived from the ET-XGB model.

The scatter plots in Figure 2 further illustrate the correlation between the actual and predicted values for the three strength properties. The data points from both the training and testing sets are closely organized around the ideal prediction line, with the majority of values residing within the ±10% error range. The consistency seen across the datasets demonstrates the accuracy of the ET-XGB model in predicting mechanical strength parameters.

The comparative analysis of different statistical indicators for the training and testing sets is presented in Figure 3.  The constantly high $R^2$ and Pearson correlation coefficients, along with minimal RSE and RRMSE values, confirm the superior prediction accuracy of the ET-XGB model. In addition, the performance index (PI) values, however relatively small, highlight the model's ability to maintain predictive stability across different datasets.

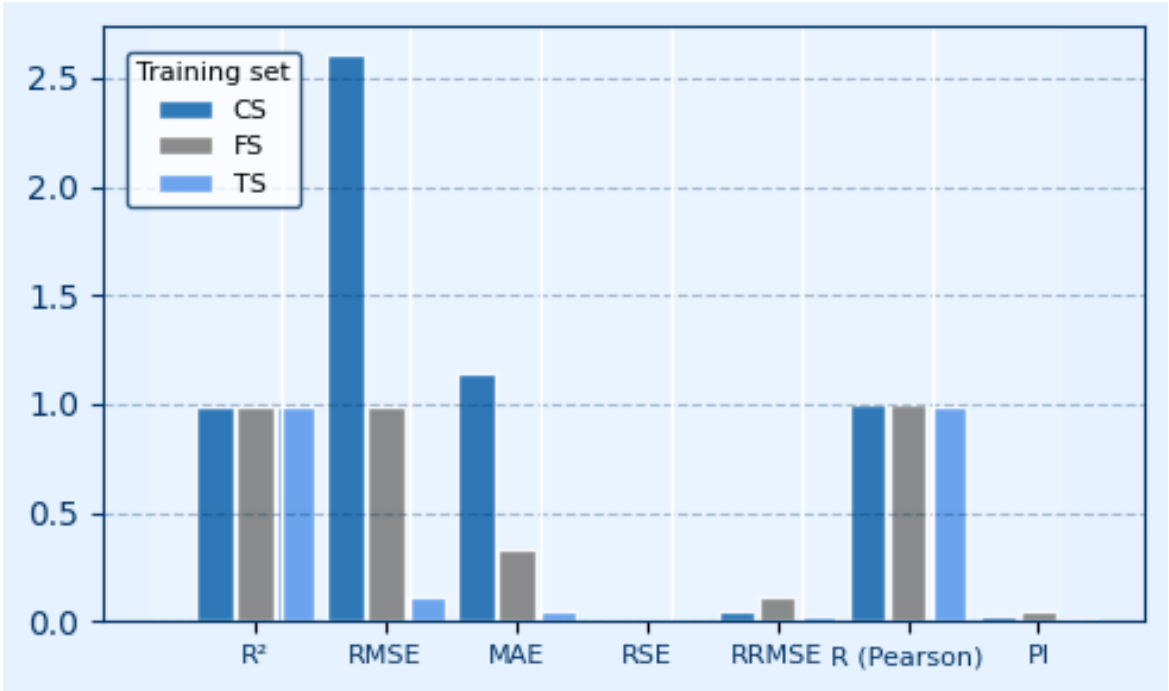

3(a)

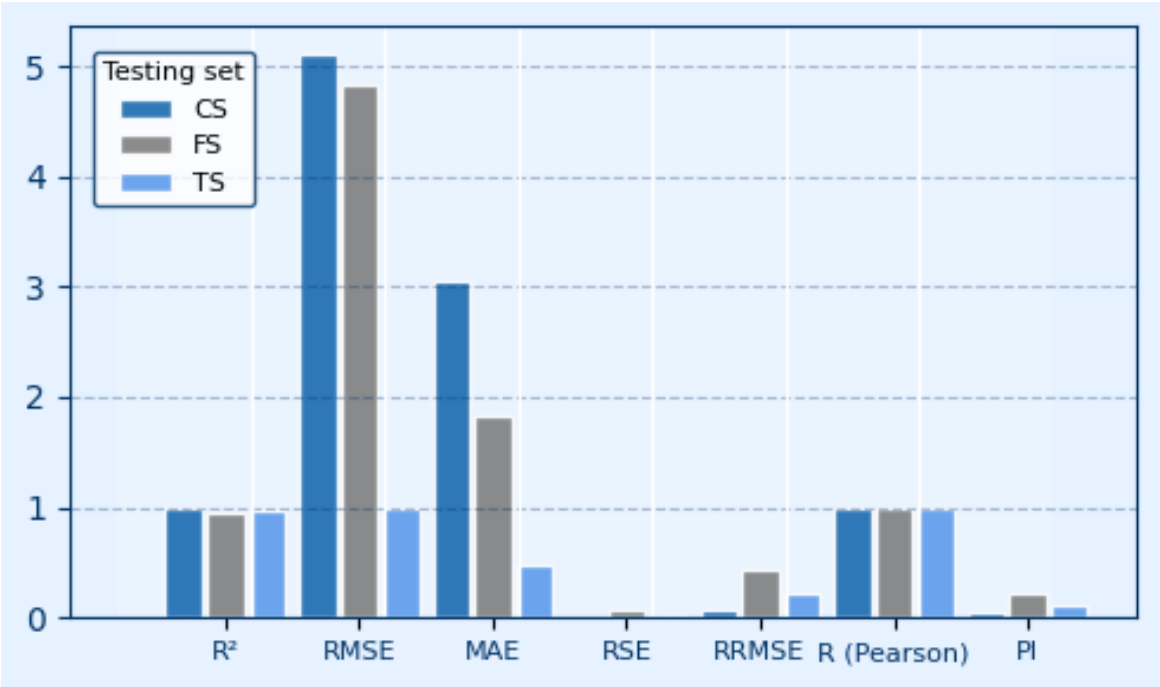

3(b)

Fig.3. Performance of ET-XGB algorithm: (a) Training, and (b) Testing

The ET-XGB hybrid model demonstrated exceptional predictive power for CS, FS, and TS, with evident strong generalization and resilience. Its performance exceeds that of traditional standalone models by utilizing the synergistic advantages of Extra Trees for variance reduction and XGB for enhancing weak learners. This combination enables ET-XGB to serve as a powerful tool for precisely predicting the mechanical properties of advanced cementitious composites.

## 3.2 Random Forest-LGBM model

The Random Forest–LGBM (RF-LGBM) hybrid model was constructed to integrate the robustness of Random Forest (RF) with the gradient boosting efficacy of LightGBM (LGBM), with the objective of enhancing the prediction of compressive strength (CS), flexural strength (FS), and tensile strength (TS). The optimized hyperparameters for both components are listed in Table 6. For the RF model, the number of estimators was established at 50, with a minimum sample split of 2 and a minimum sample leaf of 1, while the LGBM component was tuned with a learning rate of 0.005, a maximum depth of 3, and regularization parameters ($\alpha$ = 5.0, $\lambda$ = 5.0). These meticulously selected values ensured a balance between computational efficiency and predictive accuracy.

Table 6. The optimal values of the RF-LGBM model for CS, FS, and TS.

| **Hybrid algorithms** | **Hyperparameter optimization** | | **Optimal values** |
|---|---|---|---|
| (2) **Random forest - LGBM** | **RF** | No. of estimators | 50 |
| | | Max depth | None |
| | | Min samples leaf | 1 |
| | | Min sample split | 2 |
| | | Random state | 42 |
| | | No of jobs | -1 |
| | **LGBM** | No. estimators | 50 |
| | | Learning rate | 0.005 |
| | | Max depth | 3 |
| | | Min child weight | 10 |
| | | No of jobs | -1 |
| | | Reg alpha | 5.0 |
| | | Reg lambda | 5.0 |
| | | Random state | 42 |

Table 7. Evaluation metrics of the RF-LGBM model for CS, FS, and TS.

| **RF-LGBM** | **Training set** | | | **Testing set** | | |
|---|---|---|---|---|---|---|
| | CS | FS | TS | CS | FS | TS |
| **$R^2$** | 0.992 | 0.980 | 0.996 | 0.976 | 0.977 | 0.912 |
| **RMSE** | 3.990 | 2.097 | 0.561 | 7.399 | 2.100 | 0.538 |
| **MAE** | 2.163 | 0.744 | 0.209 | 4.631 | 1.031 | 0.362 |
| **RSE** | 0.008 | 0.019 | 0.003 | 0.024 | 0.022 | 0.088 |
| **RRMSE** | 0.073 | 0.224 | 0.106 | 0.129 | 0.209 | 0.145 |
| **R (Pearson)** | 0.996 | 0.997 | 0.999 | 0.988 | 0.991 | 0.955 |
| **PI** | 0.037 | 0.112 | 0.053 | 0.065 | 0.105 | 0.074 |

The performance metrics of the RF-LGBM model are displayed in Table 7. On the training set, the model achieved high $R^2$ values of 0.992, 0.980, and 0.996 for CS, FS, and TS, respectively. The RMSE values were comparatively low, and the MAE values demonstrated acceptable error levels for all strength estimates. Pearson correlation coefficients (R) exceeded 0.99 in every instance, confirming strong linear agreement between the predicted and actual values.

On the testing set, the RF-LGBM model maintained strong performance, achieving $R^2$ values of 0.976 for CS, 0.977 for FS, and 0.912 for TS. Although predictions for compressive and flexural strength exhibited good accuracy with acceptable error limits, the prediction for tensile strength had diminished generalization, as indicated by higher RMSE (7.399 for CS, 2.100 for FS, and

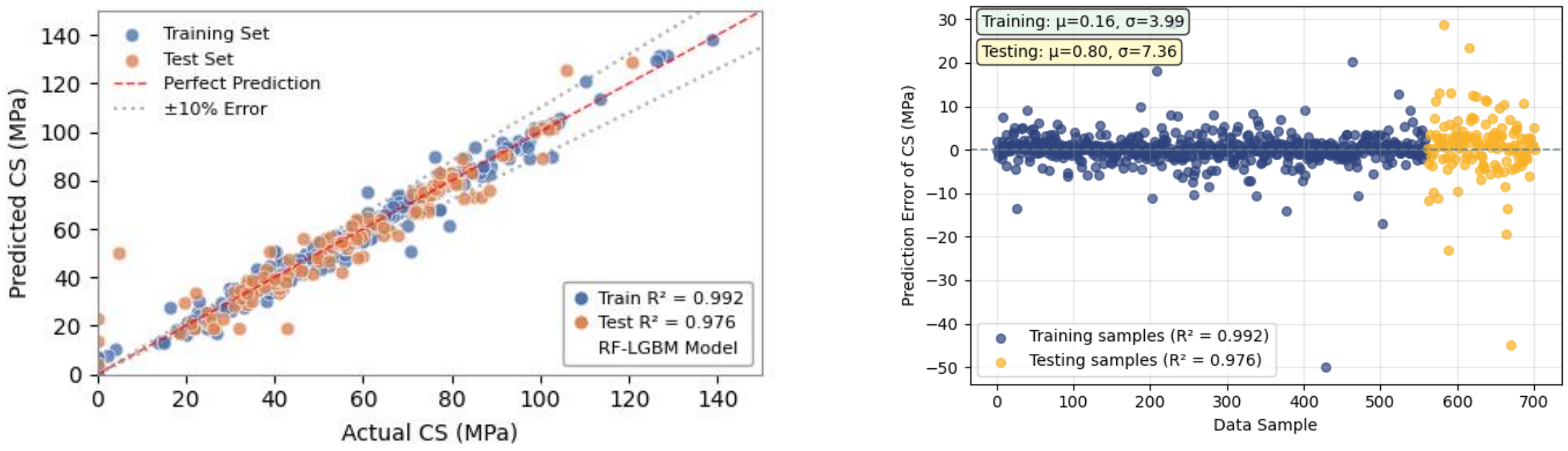

4(a)

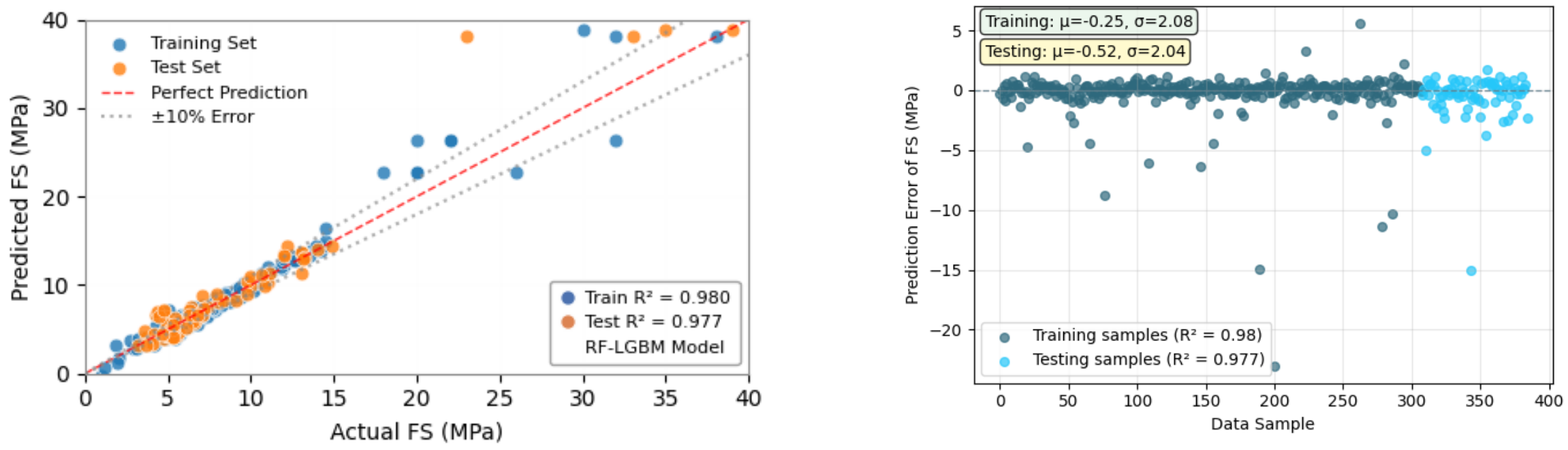

4(b)

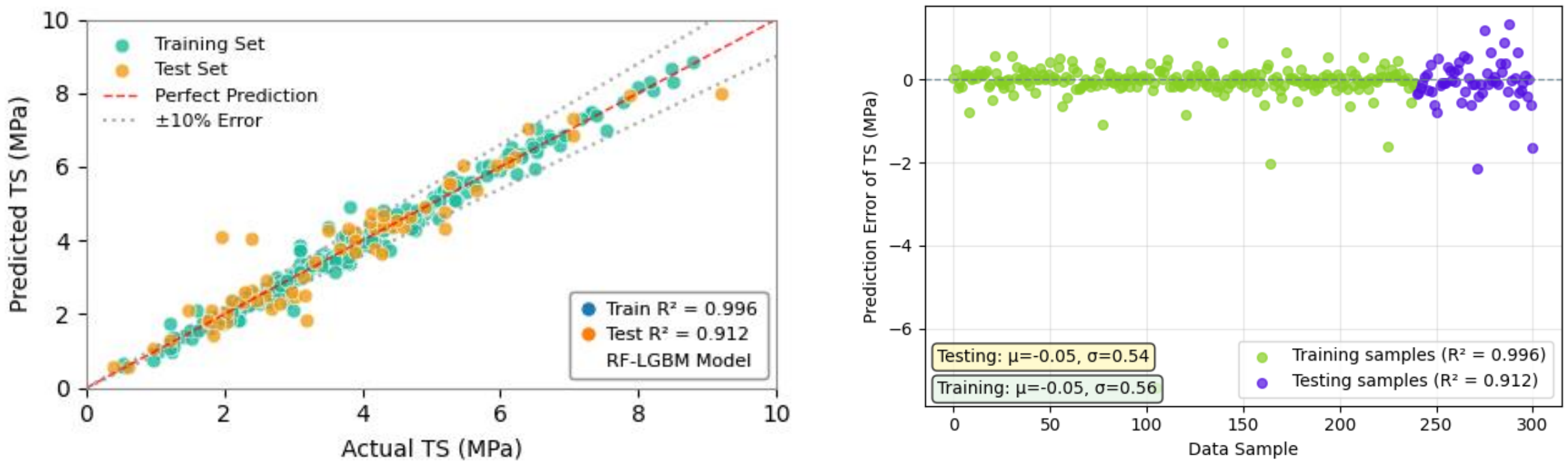

4(c)

Fig. 4. illustrates the correlation with actual and predicted results for (a) CS, (b) FS, and (c) TS derived from the RF-LGBM model.

0.538 for TS) and MAE values in compared with the training set. However, the Pearson correlation coefficients above 0.95, indicating that the model-maintained stability with unseen data.

The scatter plots in Figure 4 illustrate the correlation between the actual and predicted values for CS, FS, and TS. The predicted values are generally organized along the perfect prediction line, with the majority of data points residing within the ±10% error zone. Nevertheless, a slightly wider dispersion is observed in the TS predictions in comparison to CS and FS, which corresponds with the comparatively lower $R^2$ value.

The comparative analysis of statistical metrics (Figure 5) validates these observations. Both the training and testing datasets exhibited consistently high $R^2$ and Pearson correlation coefficients across all properties, although RMSE and MAE values indicated significant discrepancies, particularly concerning tensile strength. The performance index (PI) values further validated the reliability of the RF-LGBM model, yet they were inferior to the ET-XGB results.

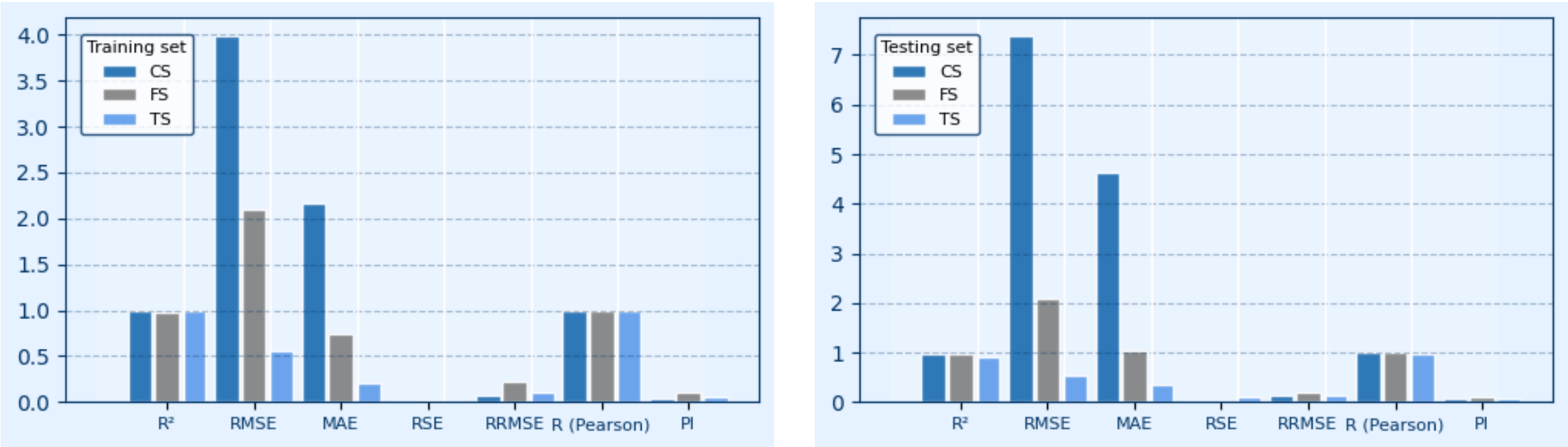

5(a) 5(b)

Fig.5. Performance of RF-LGBM algorithm: (a) Training, and (b) Testing

The RF-LGBM model exhibited strong predicted accuracy for compressive and flexural strength, albeit with slightly diminished performance for tensile strength. The hybridization of RF and LGBM improves generalization ability, enabling this method an efficient and computationally practical tool for predicting mechanical properties, though with some limitations in TS prediction accuracy.

3.3 Transformer-XGB model

The Transformer–XGB hybrid model was developed to use the sequence-learning ability of the Transformer architecture with the boosting performance of the XGB framework. The optimal hyperparameters for the Transformer component comprised two encoder layers, four attention heads, a hidden size of 128, a dropout rate of 0.1, and the Adam optimizer with a learning rate of 0.001. The model was trained for a maximum of 100 epochs, employing early stopping contingent upon validation loss. The XGB component utilized 500 estimators, a learning rate of 0.01, a maximum depth of 10, subsample and column sample ratios of 0.8, and regularization parameters (α = 1.0, λ = 1.0), as detailed in Table 8. These hyperparameters enabled the hybrid model's achieving efficient representation learning and resilient nonlinear prediction.

Table 8. The optimal values of the Transformer-XGB model for CS, FS, and TS.

| Hybrid algorithms | Hyperparameter optimization | | Optimal values |
|---|---|---|---|
| (3) **Transformer - XGB** | **Transformer (Architecture)** | No. of encoder layers | 2 |
| | | No. of attention heads | 4 |
| | | Feed-forward dimension (Hidden size) | 128 |
| | | Dropout rate | 0.1 |
| | | Optimizer | Adam |
| | | Learning rate | 0.001 |
| | | Loss function | MSE |
| | | Batch size | 32 |
| | | No. of epochs | 100 (early stops 10) |
| | | Early stopping criterion | Validation loss |
| | **XGB** | No. of estimators | 500 |
| | | Learning rate | 0.01 |
| | | Subsample | 0.8 |
| | | Max depth | 10 |
| | | Reg alpha | 1.0 |
| | | Reg lamda | 1.0 |
| | | Random state | 42 |
| | | No. of jobs | -1 |
| | | Col sample by tree | 0.8 |

Table 9. Evaluation metrics of the RF-LGBM model for CS, FS, and TS.

| **Transformer -XGB** | **Training set** | | | **Testing set** | | |
|---|---|---|---|---|---|---|
| | CS | FS | TS | CS | FS | TS |
| **$R^2$** | 0.993 | 0.994 | 0.995 | 0.981 | 0.967 | 0.978 |
| **RMSE** | 3.634 | 1.125 | 0.602 | 6.598 | 2.554 | 0.992 |
| **MAE** | 1.741 | 0.529 | 0.188 | 3.935 | 0.996 | 0.569 |
| **RSE** | 0.006 | 0.006 | 0.004 | 0.019 | 0.033 | 0.021 |
| **RRMSE** | 0.066 | 0.121 | 1.118 | 0.115 | 0.255 | 0.216 |
| **R (Pearson)** | 0.998 | 0.998 | 0.999 | 0.991 | 0.990 | 0.994 |
| **PI** | 0.033 | 0.060 | 0.059 | 0.058 | 0.128 | 0.109 |

The evaluation metrics of the Transformer–XGB model are presented in Table 9. The $R^2$ values for the training set were 0.993 for CS, 0.994 for FS, and 0.995 for TS, indicating strong fitting accuracy. The RMSE and MAE values were relatively low, especially for FS and TS, confirming the model's accuracy in capturing the intrinsic relationships within the training data. The Pearson correlation coefficients (R) reached 0.99 for all properties, highlighting the robust correlation between predicted and actual values.

In the testing set, the model demonstrated excellent predictive capability, with $R^2$ values of 0.981 for CS, 0.967 for FS, and 0.978 for TS. Although the compressive strength (CS) prediction indicating a slightly higher RMSE (6.598) and MAE (3.935) compared to the training set, the overall accuracy remained strong. Predictions of flexural and tensile strength exhibited robust generalization, with $R^2$ values exceeding 0.96 and minimal error metrics. The Pearson correlation coefficients varied between 0.990 and 0.994, verifying the model's consistency across datasets.

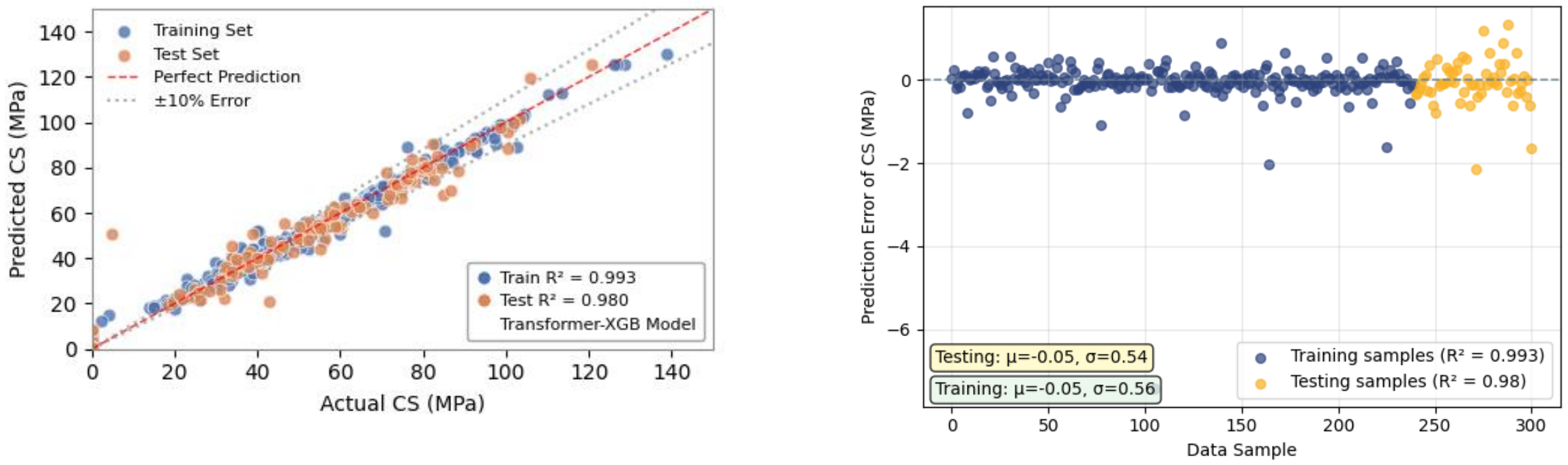

6(a)

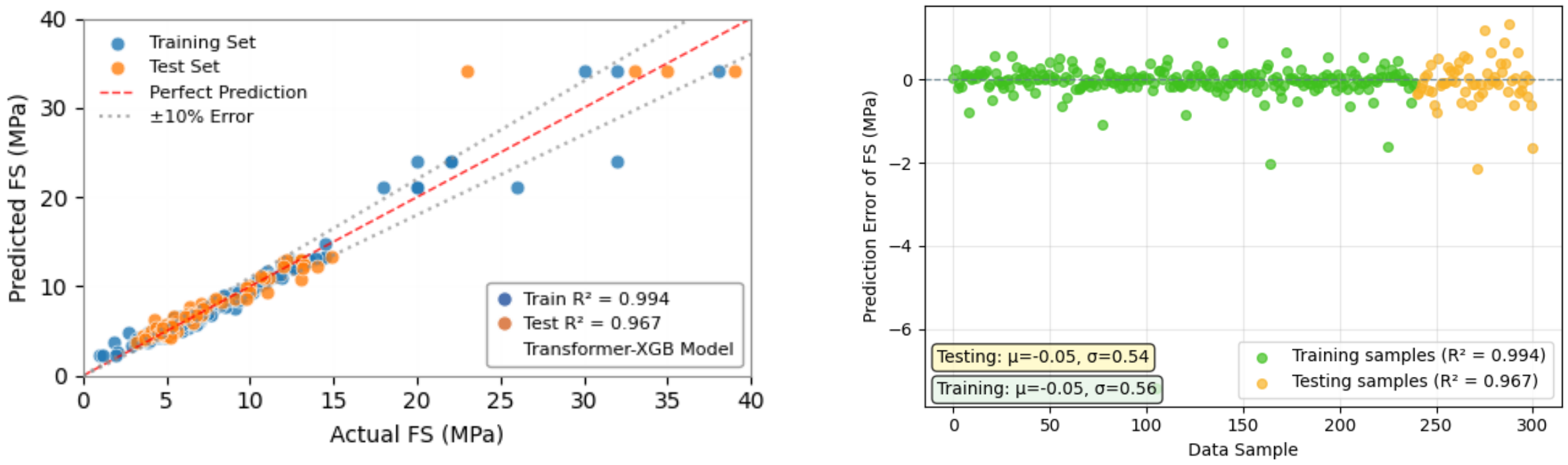

6(b)

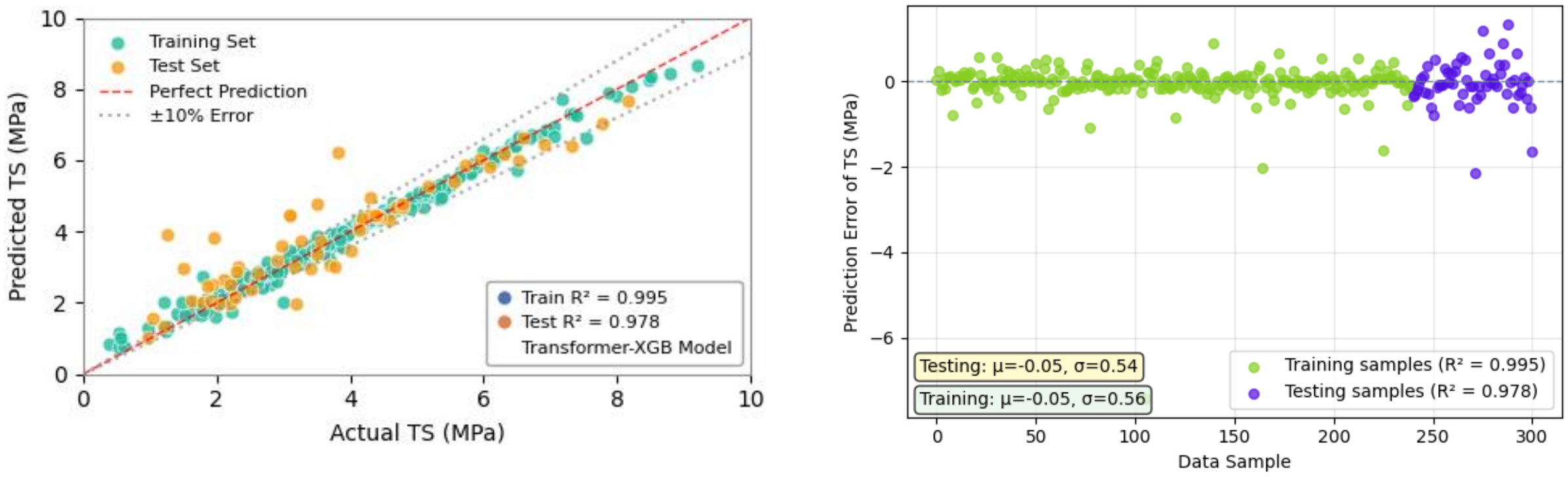

6(c)

Fig. 6. illustrates the correlation with actual and predicted results for (a) CS, (b) FS, and (c) TS derived from the Transformer-XGB model.

The correlation between the actual and predicted values of CS, FS, and TS is shown in the scatter plots in Figure 6. Data points from both the training and testing sets closely aligned with the perfect prediction line, and most predictions resided within the ±10% error variance. This indicates the Transformer–XGB model's ability to effectively capture nonlinear correlations and provide accurate predictions across different mechanical properties.

The performance metrics presented in Figure 7 further corresponds the results. The model consistently obtained high $R^2$ and Pearson correlation coefficients for all properties across both training and testing datasets. Although the slightly higher RMSE and MAE values for compressive strength in comparison to flexural and tensile strength, the relative errors (RSE and RRMSE) remained low, indicating acceptable error ranges. The performance index (PI) also validated consistent predictive capabilities across different datasets.

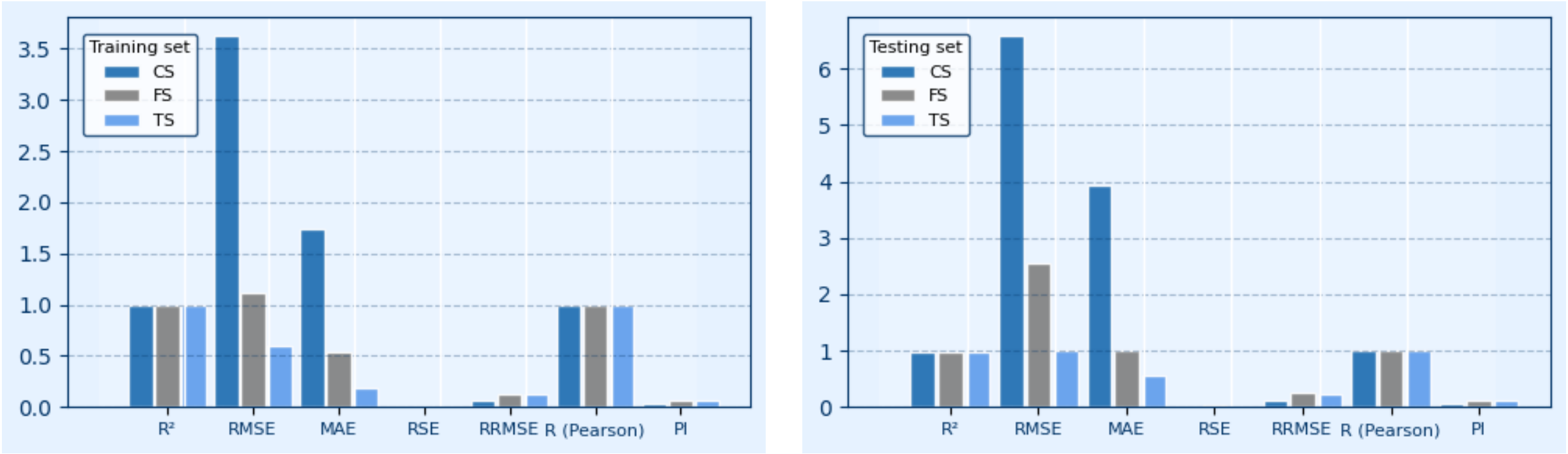

7(a) 7(b)

Fig.7. Performance of Transformer-XGB algorithm: (a) Training, and (b) Testing

The Transformer–XGB model exhibited excellent predicted accuracy for compressive, flexural, and tensile strength, with high accuracy and generalization capability. This hybrid approach, which integrates the attention-based sequence modeling of the Transformer with the gradient boosting approach of XGB, has proven efficient and reliability in predicting the mechanical properties of cementitious composites.

## 4. Uncertainty analysis of hybrid algorithms

This study was used two uncertainty analysis equations to assess the generalizability of three hybrid machine learning models regarding compressive strength, flexural strength, and tensile strength. These equations, presented earlier as Eq. (13) and Eq. (14), assess model uncertainty in

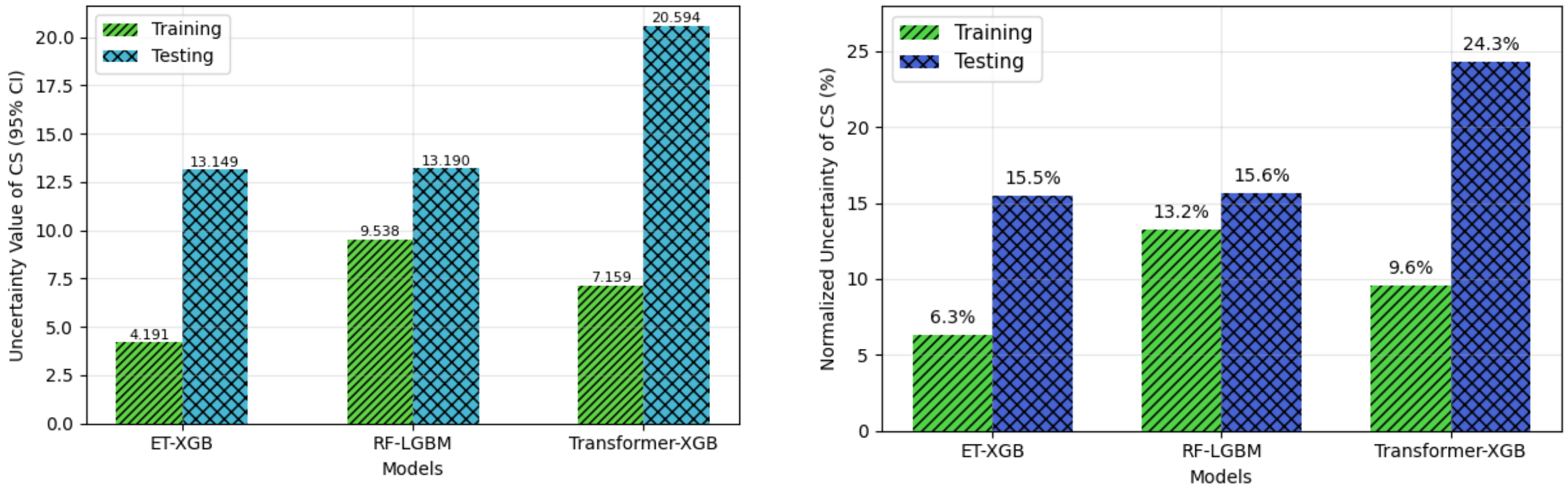

8(a) 8(b)

Fig. 8. Uncertainty analysis comparison for CS: Equation 13 (95% CI) vs. Equation 14 (Normalized % Uncertainty).

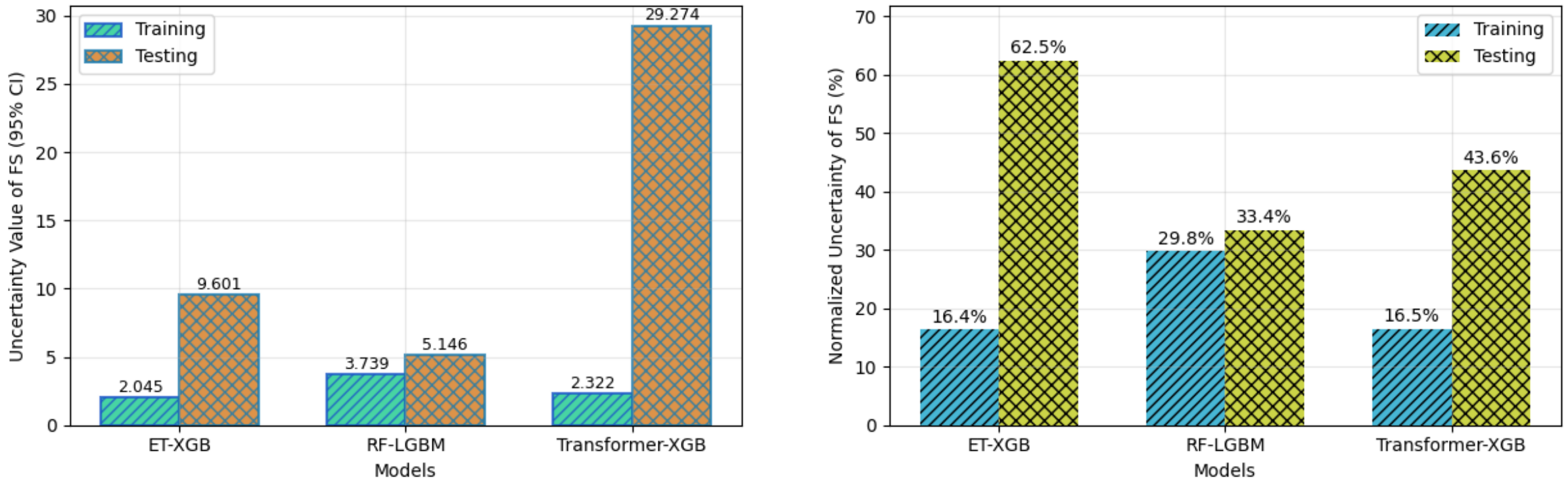

9(a) 9(b)

Fig. 9. Uncertainty analysis comparison for FS: Equation 13 (95% CI) vs. Equation 14 (Normalized % Uncertainty).

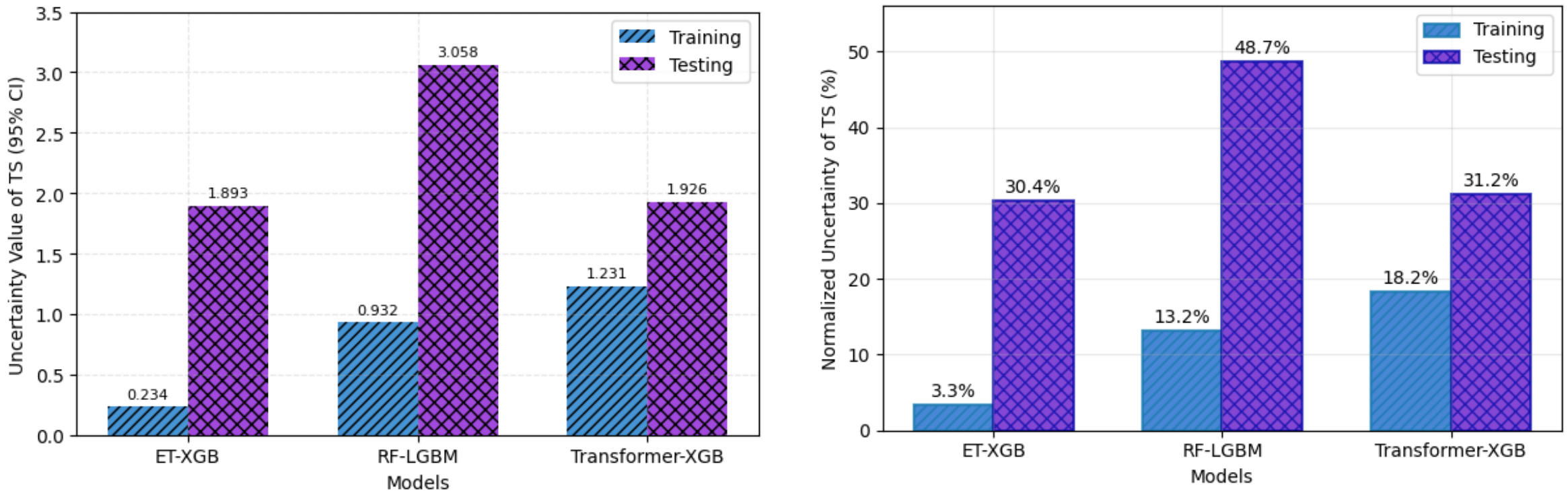

10(a) 10(b)

Fig. 10. Uncertainty analysis comparison for TS: Equation 13 (95% CI) vs. Equation 14 (Normalized % Uncertainty).

both absolute and normalized forms, thereby enabling a comprehensive comparison of predictive reliability. The uncertainty related to compressive strength (CS) prediction was evaluated using the absolute CI95% index and the normalized percentage uncertainty. As shown in Fig. 8, Transformer–XGB exhibits the largest testing uncertainty (20.59% and 24.3%), signifying inadequate generalization and significant predictive variability. RF–LGBM demonstrates comparatively high uncertainty during testing (13.19% and 15.6%). On the other hand, ET–XGB attains the lowest uncertainty compared to all models, exhibiting testing CI95% and normalized uncertainty values of 13.15% and 15.5%, respectively. Despite ET–XGB exhibiting a noticeable rise in uncertainty from training to testing, it continues to be the most stable and dependable model for CS. These results indicate that ET–XGB delivers the most reliable estimates of compressive strength, surpassing both RF–LGBM and Transformer–XGB.

The uncertainty for flexural strength (FS) was assessed utilising the identical two criteria. As illustrated in the figure. 9, Transformer–XGB exhibits the highest testing uncertainty (29.27 and 43.6%), indicating considerable instability and poor generalisation. ET–XGB has significant normalised uncertainty throughout testing (62.5%), reflecting substantial sensitivity to novel data. In contrast, RF–LGBM consistently exhibits the lowest and most stable uncertainty, with testing CI95% and normalised uncertainty values of 5.15 and 33.4%, respectively. Moreover, RF–LGBM has a reduced variance across training and testing uncertainties relative to the other models, indicating superior robustness. Consequently, both uncertainty measures identify RF–LGBM as the most dependable model for FS, above ET–XGB and Transformer–XGB.

The uncertainty associated with tensile strength (TS) prediction was assessed using CI95% and normalized percentage uncertainty, as illustrated in Fig. 10. RF–LGBM exhibits the highest testing uncertainty (3.06 and 48.7%), signifying diminished reliability. Transformer–XGB and ET–XGB have markedly reduced uncertainty levels, with ET–XGB attaining the lowest testing CI95% (1.89) and normalized uncertainty (30.4%). While Transformer–XGB has comparable performance (1.93 and 31.2%), ET–XGB exhibits the least overall uncertainty and a diminished variance between training and testing values. This consistency demonstrates robust generalization behavior. According to both indices, ET–XGB delivers the most reliable predictions for TS, surpassing RF–LGBM and Transformer–XGB in predictive stability and uncertainty control.

## 5. SHAP-based feature interpretation

Based on the uncertainty analysis, the ET–XGB model exhibited the lowest and most consistent uncertainty for compressive strength (CS), and tensile strength (TS). RF-LGBM demonstrated the lowest uncertainty, higher accuracy for flexural strength (FS). Accordingly, these two hybrid models were chosen for SHAP analysis to elucidate the contribution of input variables to the predictions of CS, FS, and TS.

Figure 11 demonstrates the feature importance and SHAP summary plots for CS using the ET–XGB model. The most influential features are PPF fiber aspect ratio (AR2), silica fume (Sfu), steel fiber content (SF), and superplasticizer (SP). On the other hand, specimen length and

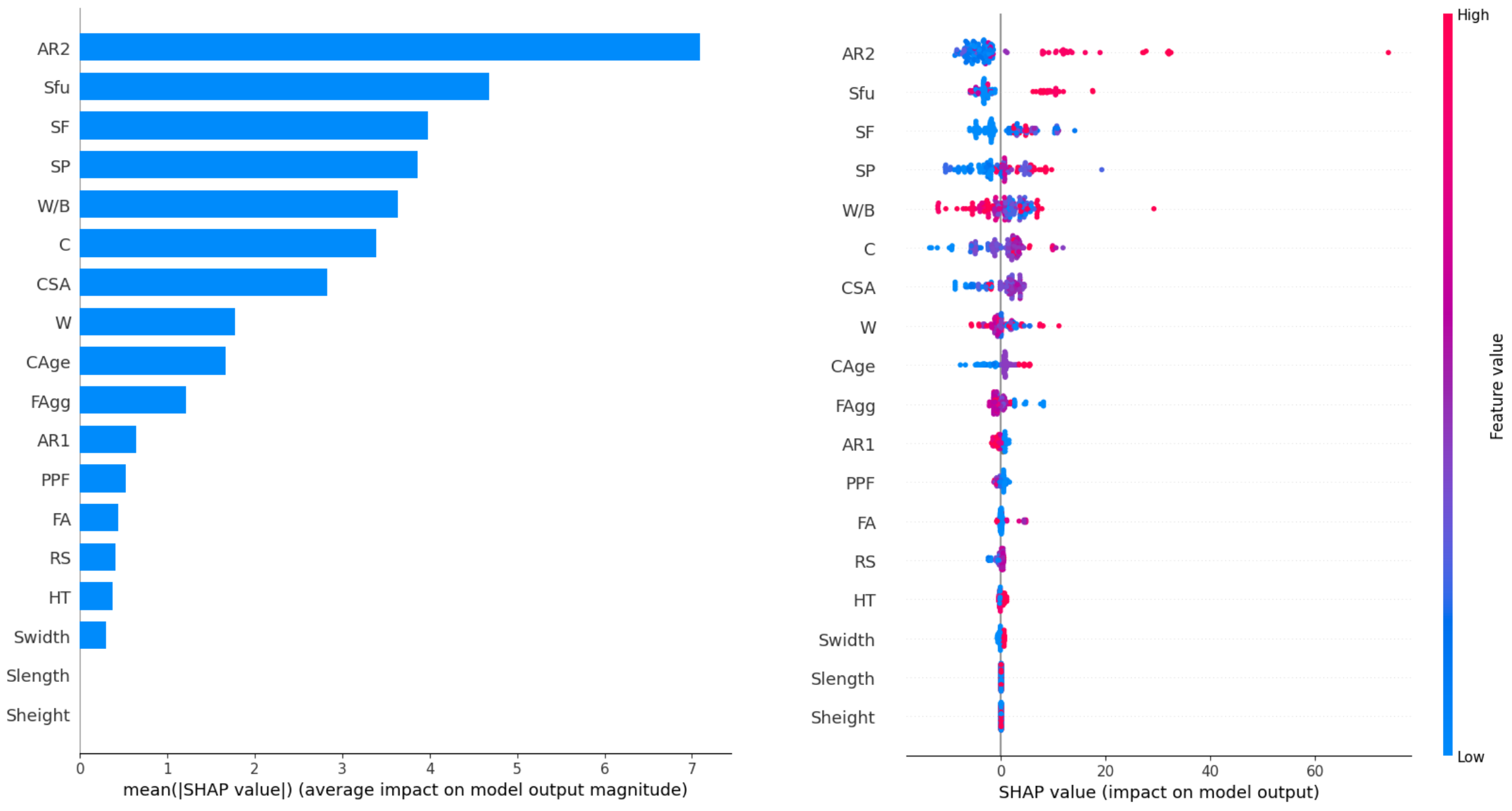

11(a) 11(b)

Fig. 11. Feature Importance and SHAP values plot for CS, ET-XGB model

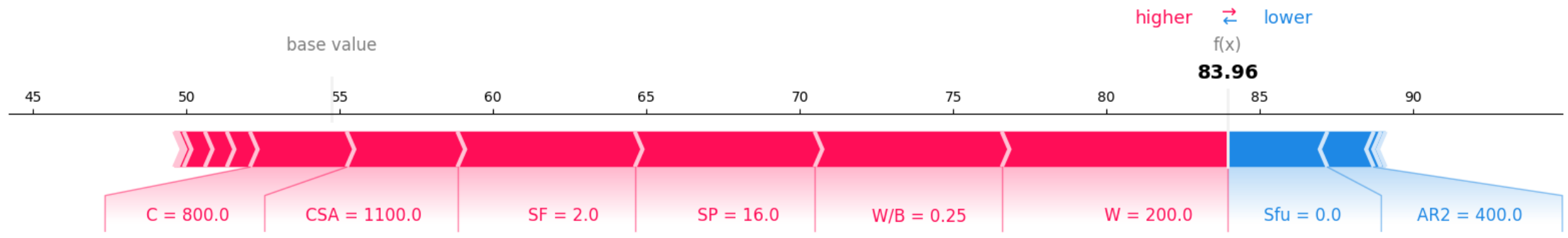

Fig. 12. SHAP values of the individual sample plot for CS, ET-XGB model

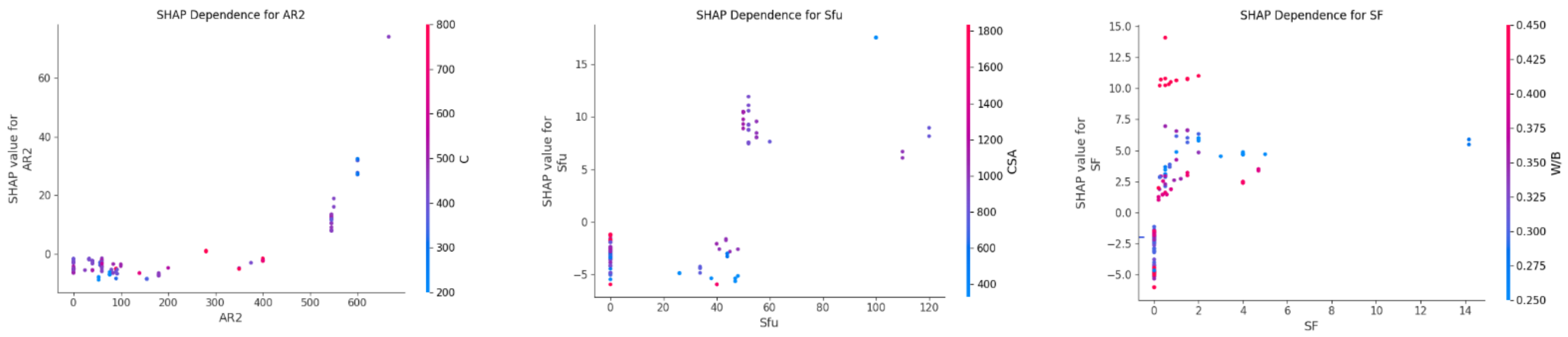

13(a) 13(b) 13(c)

Fig. 13. SHAP partial dependency plots for the CS, ET-XGB model

specimen height contributed insignificantly. The SHAP summary figure suggests that elevated AR2 and Sfu values enhance CS prediction, however a higher water–binder ratio (W/B) has a negative influence.

Figure 12 presents the SHAP values for a specific sample. In this scenario, high cement content (C = 800), substantial CSA = 1100, and moderate SF = 2.0 jointly enhanced the predicted CS, however high AR2 = 400 and low Sfu = 0 negatively impacted the prediction, diminishing it. Figure 13 displays SHAP dependency plots for the three most significant features. CS has a distinct positive association with AR2 and Sfu, indicating that increased PPF fiber aspect ratios and elevated silica fume levels improve compressive strength. Likewise, SF has beneficial effects, but with decreasing returns as its dose escalates. In contrast, elevated W/B ratios are often linked to reduced CS, in accordance with standard material behavior. The SHAP analysis indicates that AR2, Sfu, SF, and SP are the primary characteristics influencing compressive strength in ET–XGB, but high binder ratios diminish the predicted values.

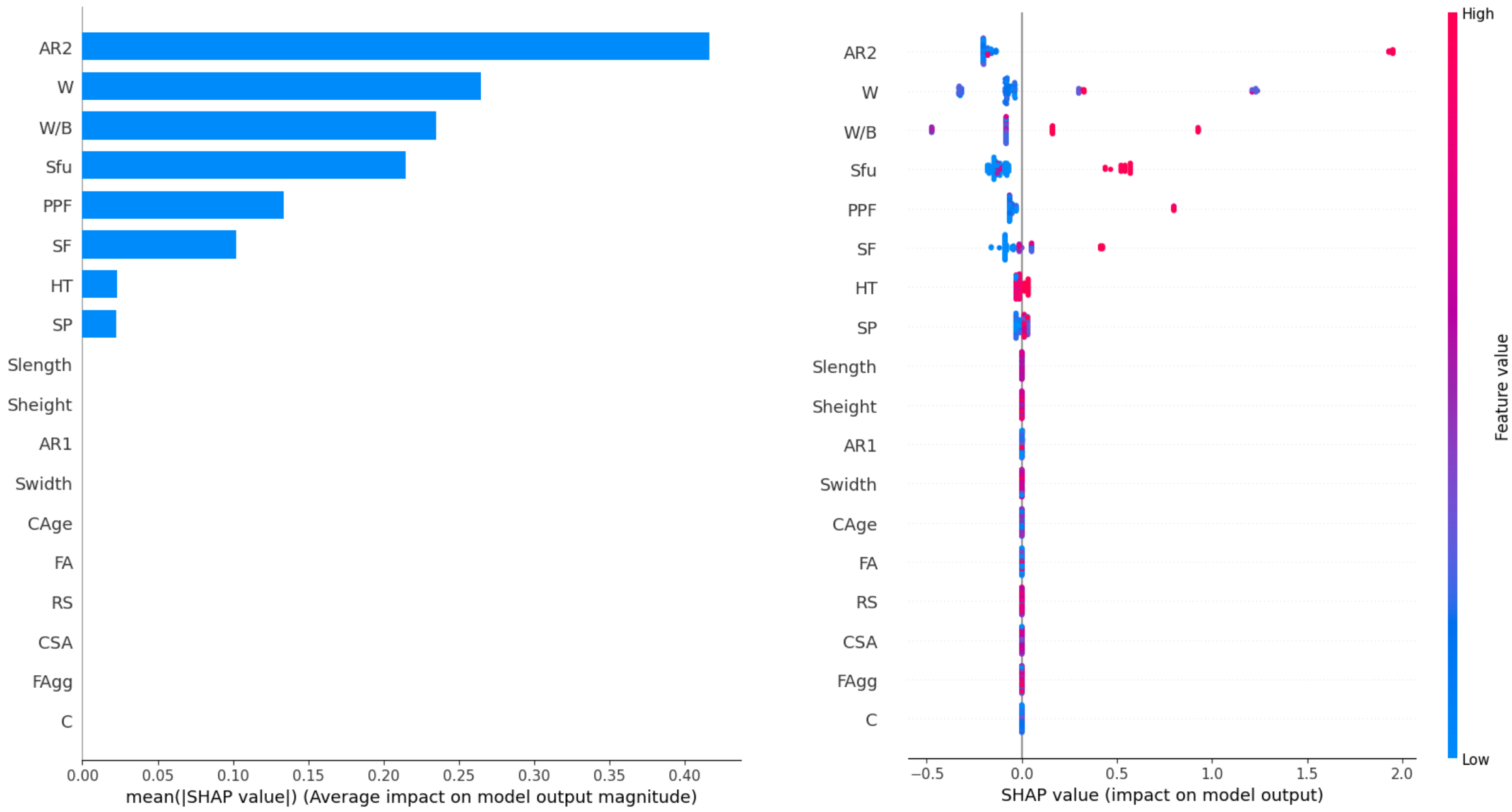

14(a) 14(b)

Fig. 14. Feature Importance and SHAP values plot for FS, RF-LGBM model

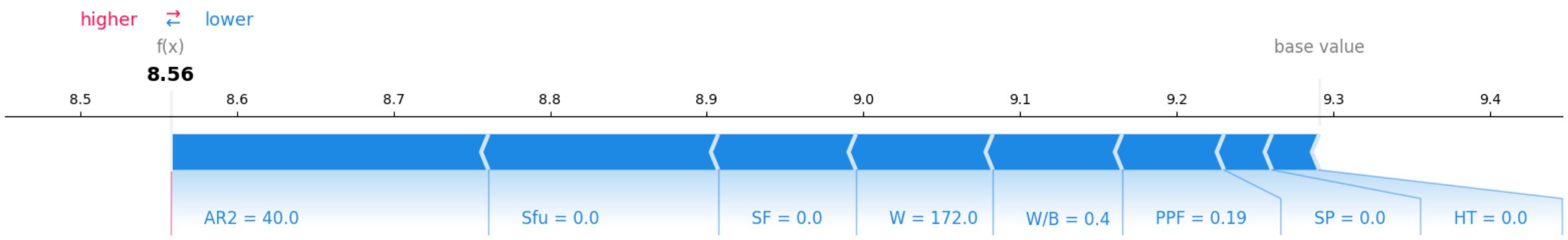

Fig. 15. SHAP values of the individual sample plot for FS, RF-LGBM model

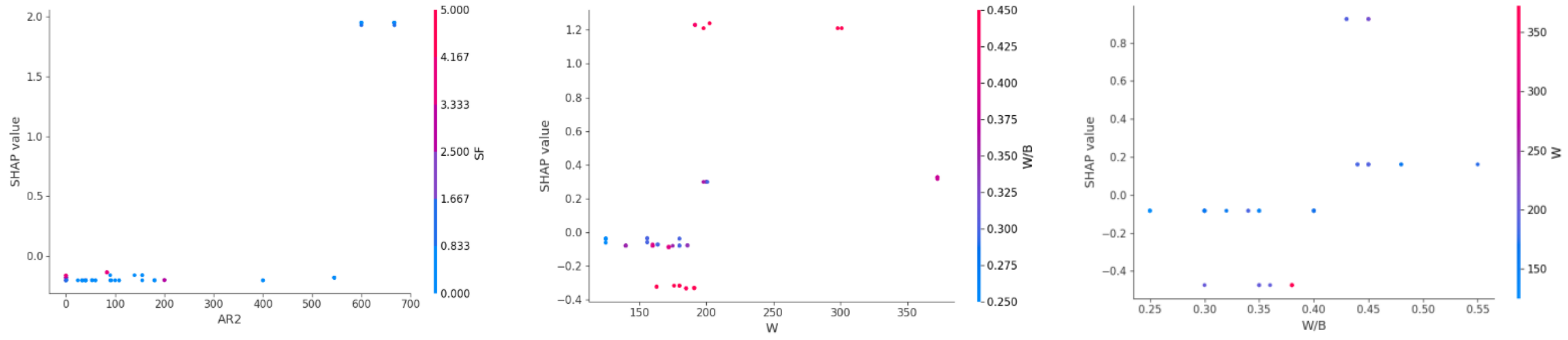

16(a) 16(b) 16(c)

Fig. 16. SHAP partial dependency plots for the FS, RF-LGBM model

Figure 14 presents the feature importance and SHAP summary plots for FS using the RF–LGBM model. The most important features are the PPF fiber aspect ratio (AR2), water content (W), water–binder ratio (W/B), and silica fume (Sfu), but other variables including PPF dose, SF, and SP exerted smaller effects. The SHAP summary plot demonstrates that elevated AR2 and Sfu values enhance FS, whereas increased W and W/B ratios diminish it.

Figure 15 illustrates the SHAP values for a specific sample. An elevated AR2 of 40.0 and a lower W of 172 enhanced FS prediction, however the shortage of Sfu and SF reduced it. Minor contributions from PPF and SP were also noted.

Figure 16 shows the dependent plots of key features. FS rises with AR2, but both W and W/B exhibit negative correlations, hence substantiating the detrimental impact of excessive water content. The SHAP analysis identifies AR2, W, W/B, and Sfu as the primary features affecting FS, with AR2 and Sfu improving predictions, while water-related parameters diminish prediction.

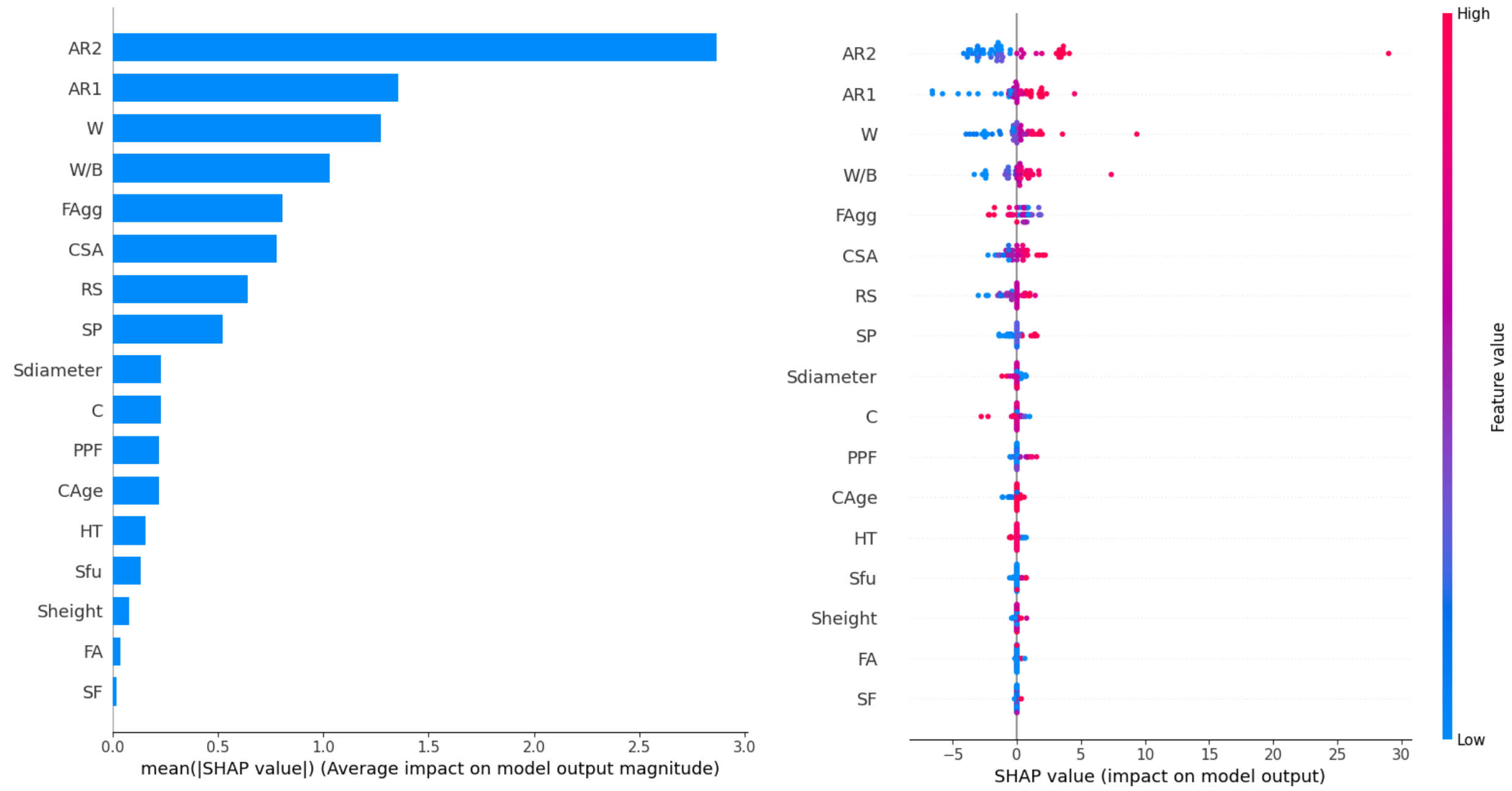

17(a) 17(b)

Fig. 17. Feature Importance and SHAP values plot for TS, ET-XGB model

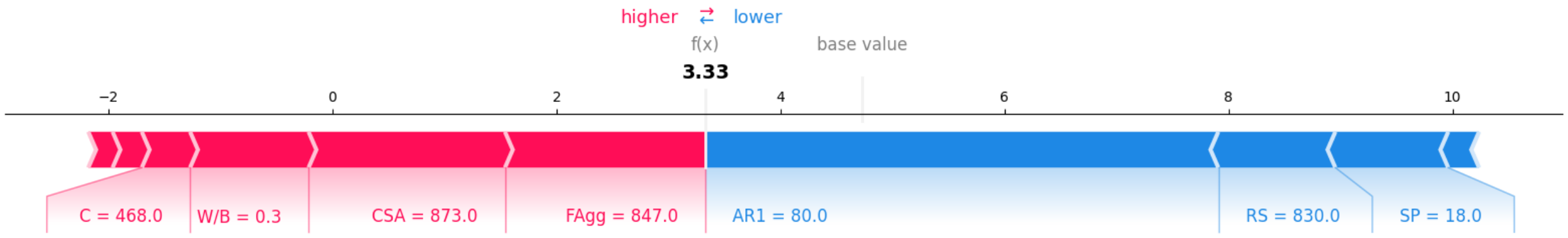

Fig. 18. SHAP values of the individual sample plot for TS, ET-XGB model

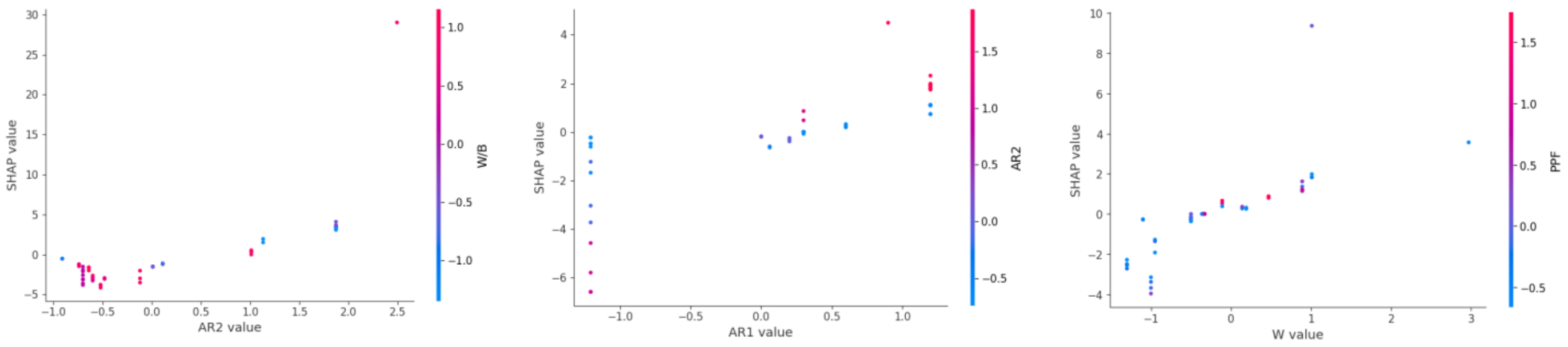

19(a) 19(b) 19(c)

Fig. 19. SHAP partial dependency plots for the TS, ET-XGB model

Figure 17 shows the feature significance and SHAP summary plots for TS using the ET–XGB model. The most influencing features are the aspect ratio of PPF fibers (AR2), the aspect ratio of steel fibers (AR1), water content (W), and the water-binder ratio (W/B). Additional parameters, including FAgg, CSA, RS, and SP, had significant influence, whilst specimen geometry and fiber doses showed negligible impact. The SHAP summary reveals that increased AR2 and AR1 values increase TS, whereas elevated W and W/B ratios diminish it.

Figure 18 illustrates the SHAP values for a specific sample. In this instance, elevated cement content (C = 468), moderate CSA = 873, and FAgg = 847 jointly enhanced TS, however low AR1 = 80 and the lack of AR2 diminished the prediction. The inclusion of RS = 830 and SP = 18 substantially enhanced the predicted TS.

Figure 19 presents SHAP dependency plots for critical properties. TS demonstrates strong positive correlations with AR2 and AR1, demonstrating the significance of fiber aspect ratios. On the contrary, both W and W/B exhibit a negative correlation, with SHAP values diminishing as these parameters rise, underscoring the adverse impact of excessive water content.

The SHAP analysis indicates that AR2, AR1, W, and W/B are the primary characteristics for predicting TS in the ET–XGB model, with fiber aspect ratios improving predictions and water-related parameters diminishing predictions.

## 6 Conclusion

This research explored three hybrid machine learning frameworks—ET–XGB, RF–LGBM, and Transformer–XGB—for predicting the compressive, flexural, and tensile strength of steel–polypropylene fiber-reinforced high-performance concrete. The models demonstrated high prediction accuracy, while uncertainty quantification evaluated their generalization reliability, and SHAP analysis produced interpretability of significant influencing features. The main findings are:

- ET–XGB: Achieved the highest overall prediction accuracy ($R^2$ = 0.994 for CS, 0.944 for FS, and 0.978 for TS) and exhibited the minimal uncertainty for both CS and TS, with testing uncertainties of roughly 13–16% for CS and 30% for TS. These findings demonstrate robust predictive stability and equitable generalization, particularly with the prediction of compressive and tensile strength.

- RF–LGBM: Performance was most consistent and dependable for flexural strength ($R^2$ = 0.977), with the lowest uncertainty for FS (approximately 5–33%).

- Transformer–XGB: Demonstrated high precision for TS ($R^2$ = 0.978) and FS ($R^2$ = 0.967), but showed the most uncertainty across all mechanical strengths (20-29% CI95% for CS and FS; 31% for TS). The significant gap between training and testing results suggests poor generalization and prediction instability.

- Uncertainty analysis (Eq. 13 vs. Eq. 14): Both absolute (95% CI) and normalized (% uncertainty) indices indicated that ET–XGB exhibited lowest uncertainty for CS and TS, while RF–LGBM was the most reliable for FS. Transformer–XGB achieved a balanced high accuracy with highest uncertainty across all strengths.
- SHAP analysis: Performed on the best- performing models identified above, SHAP showed that polypropylene fiber aspect ratio (AR2), steel fiber aspect ratio (AR1), silica fume (Sfu), and steel fiber content (SF) are significant positive contributions. In contrast, water content (W) and water–binder ratio (W/B) had negative impacts, constantly diminishing compressive strength (CS), flexural strength (FS), and tensile strength (TS) due to matrix degradation from excessive water.

Overall: ET–XGB provides the best accuracy for compressive and tensile strength, among other models. RF–LGBM ensures robust generalization for flexural strength, and Transformer–XGB achieves a balanced accuracy. Collectively, these hybrid machine learning frameworks provide potent, interpretable, and resilient tools for enhancing high-performance computing mix design and predicting structural performance.

## CRediT authorship contribution statement

**Jagaran Chakma:** Methodology, Data curation, Data generation, Formal analysis, Validation, Writing–original draft, Investigation. **Zhiguang Zhou:** Supervision, Conceptualization, Investigation, Methodology, Writing – review & editing. **Badhan Chakma**: Data generation, Data curation, Methodology, Visualization, Writing – review & editing.

## Declaration of competing interest

The authors declare that they have no known competing financial interests or personal relationships that could have appeared to influence the work reported in this paper.

## Data availability

Data will be made available on request.

## Acknowledgements

The authors wish to gratefully acknowledge the support of this work by the National Key Research and Development Program of China under Grant No. 2024YFC3015100.